\documentclass[a4paper,11pt]{article}
\ifdefined\pdfminorversion\pdfminorversion=7\fi
\usepackage[T1]{fontenc}
\usepackage[utf8]{inputenc}
\usepackage{lmodern}
\usepackage[margin=25mm]{geometry}
\usepackage{amsmath,amssymb}
\usepackage[numbers,sort&compress]{natbib}
\usepackage{graphicx}
\usepackage{booktabs,tabularx,multirow,array,threeparttable}
\usepackage{float,placeins,setspace}
\usepackage{ragged2e}
\usepackage{caption}
\usepackage{xurl}
\usepackage[hidelinks,unicode]{hyperref}
\hypersetup{
  pdftitle={PriCBMs: Multi-granular Concept Bottleneck Vision-Language Models for Visual Privacy Recognition},
  pdfauthor={Shenglei Pei, Haotian Sun, Pengfei Li, Bingxu Lu, Wenlong Yu},
  pdfsubject={},
  pdfkeywords={Visual privacy recognition, Concept bottleneck models, Vision-language models, Multi-label learning},
  pdfcreator={LaTeX with hyperref}
}
\newcommand{\vect}[1]{\mathbf{#1}}
\allowdisplaybreaks
\title{PriCBMs: Multi-granular Concept Bottleneck Vision-Language Models for Visual Privacy Recognition}
\newcommand{\authorblock}{\parbox{0.96\textwidth}{\centering
  \normalsize Shenglei Pei\textsuperscript{1}, Haotian Sun\textsuperscript{1},
  Pengfei Li\textsuperscript{1}, Bingxu Lu\textsuperscript{3}, Wenlong Yu\textsuperscript{2}\par
  \vspace{0.6em}
  \small
  \textsuperscript{1}School of Intelligence Science and Engineering, Qinghai Minzu University, Xining 810007, China\par
  \textsuperscript{2}School of Computing and Information Systems, Singapore Management University, Singapore 178902, Singapore\par
  \textsuperscript{3}Tianjin Key Laboratory of Interventional Brain-Computer Interface and Intelligent Rehabilitation and Academy for Advanced Interdisciplinary Studies, Nankai University, Tianjin 300350, China\par
  \vspace{0.5em}
  \footnotesize
  Emails: \href{mailto:peishenglei@126.com}{\nolinkurl{peishenglei@126.com}} (Shenglei Pei);
  \href{mailto:sun2293674491@gmail.com}{\nolinkurl{sun2293674491@gmail.com}} (Haotian Sun);\par
  \href{mailto:lipengfei@qhmu.edu.cn}{\nolinkurl{lipengfei@qhmu.edu.cn}} (Pengfei Li);
  \href{mailto:lubingxu@nankai.edu.cn}{\nolinkurl{lubingxu@nankai.edu.cn}} (Bingxu Lu);
  \href{mailto:wloongyu@gmail.com}{\nolinkurl{wloongyu@gmail.com}} (Wenlong Yu).\par
  Corresponding authors: Wenlong Yu and Bingxu Lu.\par
  Shenglei Pei ORCID: \url{https://orcid.org/0000-0002-8070-8580}
}}
\author{\authorblock}
\date{}

\begin{document}
\maketitle
\begin{abstract}

Widespread image sharing on social media has increased the need for visual privacy recognition to identify inadvertently exposed privacy information.
However, multi-label visual privacy recognition presents two challenges. First, privacy categories represent high-level semantic attributes and do not have a one-to-one correspondence with visual objects. Different visual objects may involve the same privacy category, while the same type of object may contain different types of privacy-sensitive information. In this setting, direct prediction methods generally lack explicit guidance for identifying category-relevant privacy cues, leaving this process implicit in the learned image-to-label mapping and potentially limiting classification performance.
Second, privacy cues are often distributed across image regions and may not be adequately represented by whole-image features.
To address these challenges, we propose Privacy Concept Bottleneck Models (PriCBMs). By introducing privacy-related concepts as intermediate semantic representations, PriCBMs first identify category-relevant privacy cues through visual--concept matching and then predict privacy categories based on concept responses.
This process makes privacy predictions more interpretable.
To improve the quality of these concepts, we design a concept selection strategy based on target-category relevance, cross-category discriminability, and semantic diversity.
Meanwhile, we introduce a global-region modeling module to capture localized privacy evidence while retaining global context.
Experiments on VISPR and FineViP show that PriCBMs outperform evaluated competitors across all four multi-label metrics. Ablation studies and concept-quality analyses support the effectiveness of the proposed design. Code is available at \url{https://github.com/sun2293674491-dotcom/PriCBMs}.


\end{abstract}

\section{Introduction}
\label{sec:introduction}

The rapid growth of social media and image-sharing platforms has made image sharing an increasingly common form of online communication. Images shared online may inadvertently expose sensitive information related to personal identity, location, social relationships, and other private content \cite{ref3,zhao2025visual}. This has created a growing need for effective visual privacy recognition. For visual privacy recognition, especially for screening images before online sharing, low-latency inference and on-device deployment are desirable, as cloud-based processing may introduce transmission delays and additional privacy exposure risks.

Early studies mainly formulated visual privacy recognition as a binary private/public classification problem \cite{ref1,ref36,ref21}. Binary prediction provides a basic indication of whether an image contains privacy-sensitive information. However, a private/public decision alone provides limited information about the specific privacy risks present in an image, making it difficult for users to understand and manage these risks. 
This gap between binary privacy judgments and users' need for more specific privacy information has motivated the development of multi-label visual privacy recognition. Multi-label visual privacy recognition allows multiple privacy-related labels to be identified simultaneously within the same image, thereby providing more specific information about the types of privacy-sensitive content present in the image. Existing approaches to multi-label visual privacy recognition \cite{ref3,chen2021privattnet} typically learn a direct mapping from visual features to privacy labels using multi-label classification objectives.

Despite the progress made in multi-label visual privacy recognition, two challenges remain. First, privacy categories represent high-level semantic attributes and do not have a one-to-one correspondence with visual objects. The same privacy category may involve different visual objects, while the same type of visual object may contain different types of privacy-sensitive information. This complex relationship makes it challenging to identify category-relevant privacy cues in images. Existing direct prediction methods generally lack explicit guidance for identifying category-relevant privacy cues, leaving this process implicit in the learned image-to-label mapping and potentially limiting classification performance.
Second, privacy cues are often distributed across different image regions. Although whole-image features capture overall visual information, they may not adequately represent localized privacy cues, particularly those in small or visually less prominent regions~\cite{ocb2025,tang2022pyramidal}. Relying solely on whole-image features may therefore limit the accurate and comprehensive identification of such cues.

To address these challenges, we propose Privacy Concept Bottleneck Models (PriCBMs). Based on the concept bottleneck framework \cite{ref7}, PriCBMs introduce privacy-related concepts as intermediate semantic representations between images and privacy categories. By constructing a privacy concept bank, PriCBMs match visual features with predefined privacy concepts to help identify category-relevant privacy cues across diverse visual content. Subsequently, privacy categories are predicted from concept responses. This design replaces direct image-to-label prediction with an explicit concept bottleneck that guides the model to identify category-relevant privacy cues through visual--concept matching. This structural change helps improve classification performance, while the concept bottleneck enhances model interpretability.
To improve the quality and discriminative power of the privacy concept bank, we design a privacy-oriented concept selection strategy based on three criteria, namely target-category relevance, cross-category discriminability, and semantic diversity. The selected concepts are then combined to form the privacy concept bank.
PriCBMs further introduce a global-region modeling module that enhances the capture of localized privacy cues while preserving global contextual information. Category-specific fusion weights are used to balance global context and localized evidence  for each privacy category.
The contributions of this paper are summarized as follows:
\begin{itemize}
    \item We propose Privacy Concept Bottleneck Models (PriCBMs) for multi-label visual privacy recognition. By introducing privacy-related concepts as intermediate semantic representations, PriCBMs formulate privacy prediction as an explicit image--concept--category process, thereby helping identify category-relevant privacy cues across diverse visual content.

    \item We develop a global-region modeling module that enhances the capture of localized privacy cues while preserving global contextual information. We also introduce a privacy-oriented concept selection strategy that considers target-category relevance, cross-category discriminability, and semantic diversity, yielding a compact and discriminative privacy concept bank.

    \item Experiments on two multi-label visual privacy datasets show that PriCBMs outperform the evaluated competitors across all four multi-label metrics. Ablation studies validate the proposed strategies, and concept-quality analyses confirm the quality of the selected concepts under the evaluated settings.
    
\end{itemize}

\section{Related Work}

Early studies on visual privacy recognition formulated the task as binary private/public classification. PicAlert!~\cite{ref36} combined visual features, user-provided tags, and metadata to predict whether an image was private or public. Privacy-aware image classification and search~\cite{ref1} explored privacy prediction in image classification and retrieval settings. Tonge and Caragea~\cite{ref21} investigated deep neural network representations for image privacy prediction, while Privacy-CNH~\cite{ref19} employed hierarchical visual representations for privacy classification. Yang et al.~\cite{yang2020graph} further modeled object--privacy relationships using graph-based neural networks to provide explainable image privacy inference. Some studies also considered how privacy judgments may vary across users and application scenarios. Personalized privacy-aware image classification~\cite{ref20} modeled differences in privacy preferences across users. VizWiz-Priv~\cite{ref14} investigated private visual information in images captured by blind users, while PrivacyAlert~\cite{ref37} studied image privacy prediction in social media sharing scenarios. Despite these advances, binary private/public prediction provides only a coarse assessment of privacy risk and cannot specify the types of sensitive information exposed in an image. PriCBMs therefore focus on multi-label visual privacy recognition, allowing multiple privacy categories to be identified within the same image.

Multi-label visual privacy recognition was developed to identify multiple types of sensitive information within the same image. VISPR~\cite{ref3} introduced a taxonomy of privacy attributes and established a benchmark for multi-label privacy attribute prediction. MSPP-Net~\cite{li2025mspp} adopted a multi-stage semantic perception framework that progressively modeled entities, semantic attributes, and inter-object context for privacy recognition. Related visual studies have explored label-specific~\cite{li2022common}, multi-granularity~\cite{gao2026mcfpt}, and knowledge-enhanced representations~\cite{palazzo2021structured}, including feature structuralization for fine-grained domain generalization~\cite{yu2025fine}, LLM-generated semantic descriptions for wildlife recognition~\cite{li2025wildlife}, and parameter-free knowledge embedding with fuzzy cognitive maps for class-incremental learning~\cite{wang2025spark}. These studies reflect efforts to incorporate discriminative semantic or prior knowledge into visual learning. However, in existing multi-label privacy recognition, semantic and contextual information was still mainly used to enrich learned representations, while privacy-category prediction was not explicitly constrained by category-relevant semantic evidence. As a result, models may rely on dominant or shared visual patterns rather than category-relevant cues needed to distinguish different privacy categories, potentially leading to category confusion. PriCBMs introduce category-relevant privacy concepts as intermediate representations, reformulating direct image-to-label prediction as an explicit image--concept--category process in which category prediction is mediated through privacy-concept responses.

Spatial cues have also been explored in multi-label visual privacy recognition. PrivAttNetMLC~\cite{chen2021privattnet} employed visual attention to emphasize privacy-relevant regions, while Jiang et al.~\cite{jiang2025connecting} employed privacy-specific spatial attention to capture attribute-dependent visual evidence across different image regions. These studies progressively strengthened the use of spatial cues from attention-based feature emphasis to attribute-specific spatial modeling. However, localized cues were mainly integrated through attention weighting and subsequent aggregation, which may weaken privacy evidence distributed across different or less prominent regions. PriCBMs instead explicitly model whole-image and region-level concept responses as complementary sources of privacy evidence and integrate them for multi-label category prediction across different spatial granularities. Beyond privacy recognition, Object-Centric Concept-Bottlenecks (OCB)~\cite{ocb2025} combines image- and object-level concept activations for visual classification, whereas PriCBMs use privacy-specific concepts and category-specific global-region fusion with regional features derived from pooled patch tokens.

Interpretability was also explored in multi-label visual privacy recognition. Existing approaches mainly provided spatially grounded evidence through attention or localization mechanisms, allowing users to more directly inspect where privacy-related cues occurred in an image. For example, the attention mechanism in PrivAttNet~\cite{chen2021privattnet} highlighted image regions associated with privacy predictions, while the privacy-specific attention mechanism of Jiang et al.~\cite{jiang2025connecting} exposed spatial evidence associated with individual privacy attributes. Such mechanisms improved the spatial inspectability of multi-label privacy predictions by revealing where relevant evidence was located. Beyond spatial localization, concept-based methods have also explored semantic explanations for visual predictions. CoE generates natural-language explanations by tracing visual concept circuits and analyzing concept polysemanticity~\cite{yu2025coe}. However, dedicated studies on explaining multi-label privacy predictions remain limited, particularly in providing explicit semantic evidence that clarifies what privacy-related information supports individual category predictions. In PriCBMs, the concept bottleneck layer provides an explicit semantic interface between visual representations and privacy categories. Since category predictions are mediated through privacy concepts, the concept responses and their associations with privacy categories can be inspected to understand the semantic representations associated with individual predictions, thereby providing inherent interpretability within the prediction process.

\section{Method}
\label{sec:method}

The goal of multi-label visual privacy recognition is to identify all privacy-risk categories in an image. Let \(C=C_p\cup\{c_{\mathrm{safe}}\}\), where \(C_p\) contains \(L\) privacy-risk categories and \(c_{\mathrm{safe}}\) denotes Safe/Public. For the \(j\)-th privacy-risk category \(c_j\in C_p\), the ground-truth label is \(y_j\in\{0,1\}\), where \(y_j=1\) indicates the presence of the privacy risk. The PriCBMs framework is illustrated in Fig.~\ref{fig:overall_framework}.

\begingroup
\setstretch{1}
\centering

\vspace{0.8em}

\includegraphics[
    width=1.00\linewidth,
    keepaspectratio
]{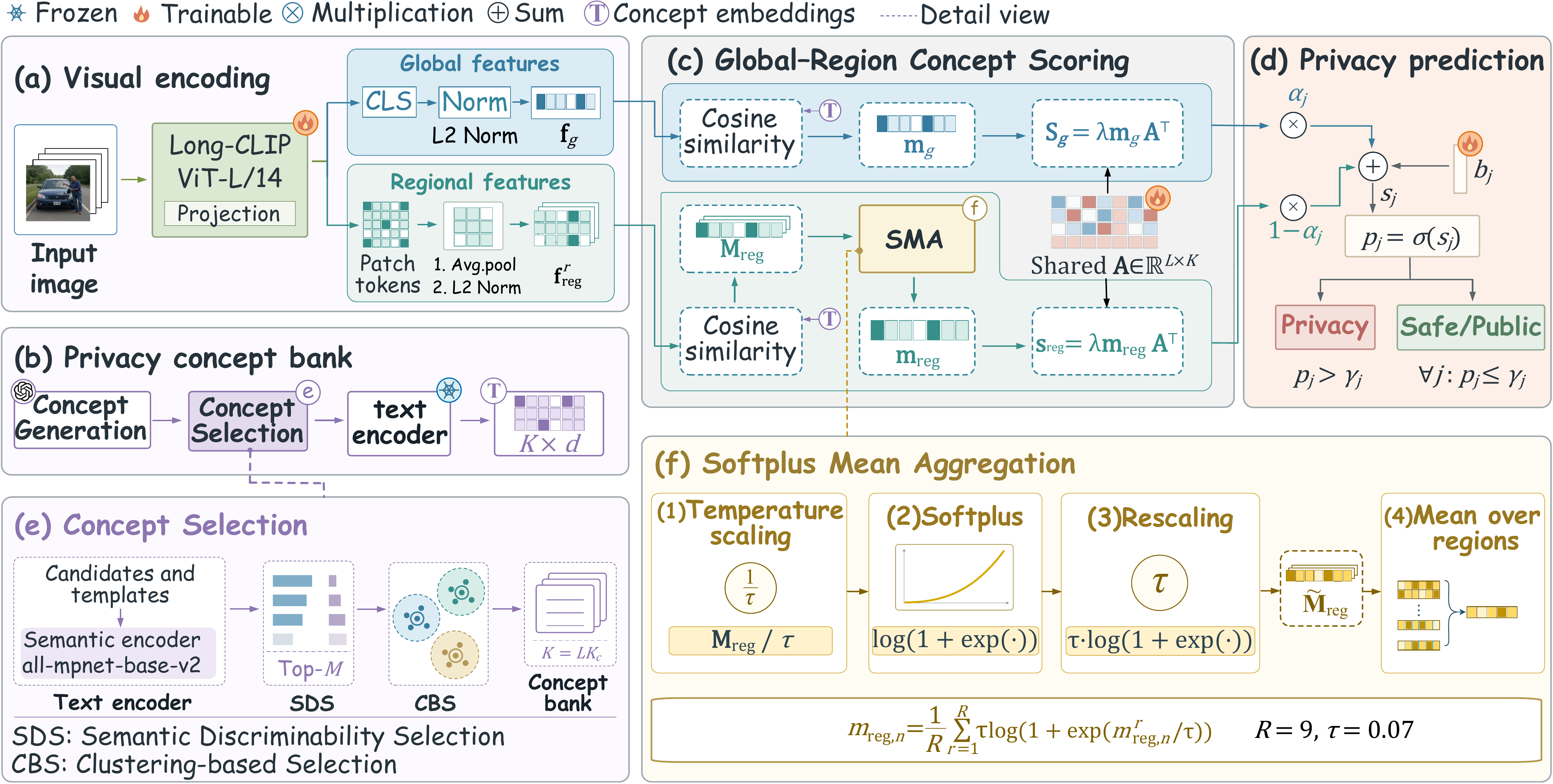}

\vspace{0.5em}
\captionof{figure}{Overall framework of the proposed PriCBMs. The example image is from VISPR~\cite{ref3}.}
\label{fig:overall_framework}

\vspace{0.8em}
\endgroup

\FloatBarrier

Unlike direct image-to-label prediction, PriCBMs introduce privacy-related natural language concepts as intermediate semantic representations, establishing an image--concept--category process. PriCBMs construct a privacy concept bank by generating category-specific candidate concepts and selecting them based on target-category relevance, cross-category discriminability, and semantic diversity.

The selected concepts and input image are encoded by Long-CLIP~\cite{zhang2024long} into a shared vision-language embedding space. The image is represented by a global feature and multiple region-level visual features.

The global image feature is matched with the concept representations to obtain global concept responses. Meanwhile, multiple regional visual features extracted from each image are matched with the concept representations to obtain concept responses for each region. These responses are aggregated using Softplus Mean Aggregation (SMA) to capture localized privacy evidence. The global and regional concept responses are mapped to category-level scores through a shared concept-to-class association matrix and adaptively fused using category-specific weights. The fused category scores are converted into multi-label prediction probabilities. 

Unlike privacy-risk categories, Safe/Public encompasses a wide range of visual content that may also appear in privacy-sensitive images, making it challenging to construct a set of discriminative visual concepts for this category. Therefore, PriCBMs treat Safe/Public as a residual category and assign it only when no privacy-risk category is activated for that image.

The following subsections detail the main components of PriCBMs and their roles in privacy recognition, beginning with privacy concept generation and selection, followed by global-region concept modeling. We then describe the inference and calibration procedure and the training objective.

\subsection{Concept Generation and Selection}
\label{subsec:concept_generation_Selection}

This section describes how PriCBMs construct the privacy concept bank through candidate concept generation, semantic discriminability selection, and clustering-based diversity selection, as illustrated in Fig.~\ref{fig:concept_selection_flowchart}.

PriCBMs use two frozen text encoders for different purposes. The all-mpnet-base-v2 encoder in the SentenceTransformer framework~\cite{reimers2019sbert}, denoted as \(E_s(\cdot)\), is used for concept selection. After concept selection, the concepts are encoded by the Long-CLIP text encoder~\cite{zhang2024long} for image--concept matching. The resulting concept embeddings are precomputed and kept fixed during training.

\begin{center}
\setstretch{1}
\includegraphics[
    width=1.00\linewidth,
    height=0.58\textheight,
    keepaspectratio
]{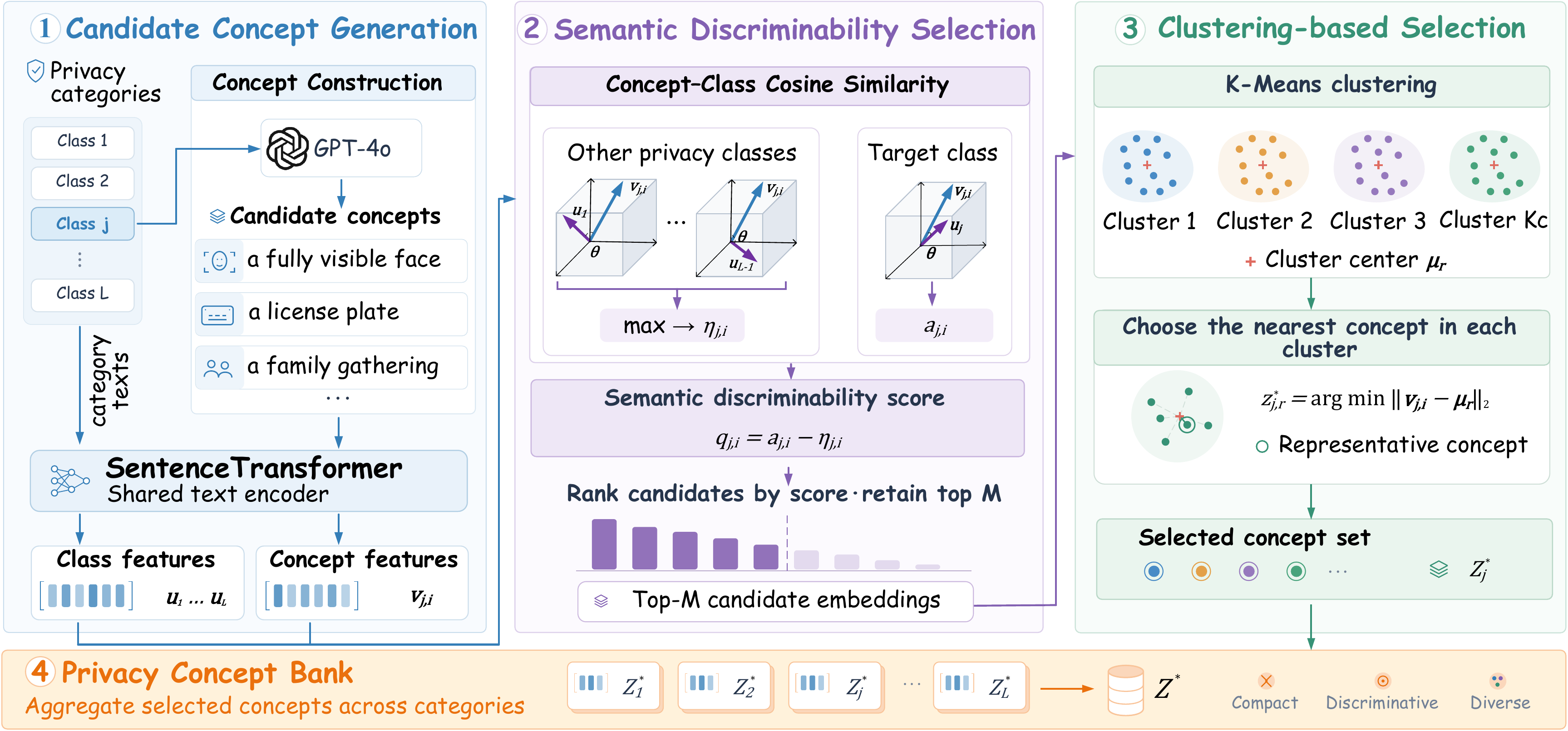}

\vspace{0.2em}
\captionof{figure}{Flowchart of the concept generation and selection process.}
\label{fig:concept_selection_flowchart}
\end{center}

\FloatBarrier

\subsubsection{Concept Construction}

For the \(j\)-th privacy-risk category \(c_j\), we prompt GPT-4o~\cite{openai2024gpt4o} to generate candidate visual concepts specifically for that category. After duplicate and invalid outputs are removed, we retain \(N\) valid candidate concepts as
\begin{equation}
Z_j=\{z_{j,1},z_{j,2},\ldots,z_{j,N}\}.
\end{equation}

\subsubsection{Semantic Discriminability Selection}

Semantic Discriminability Selection retains concepts that are relevant to the target privacy-risk category and help distinguish it from other privacy-risk categories by measuring similarity margins against other category descriptions.

Each privacy-risk category is represented by the following templated input.
\begin{equation*}
\phi(c_j)=\text{``A photo showing }c_j\text{''}.
\end{equation*}

We encode and normalize the category descriptions and candidate concepts using \(E_s(\cdot)\) as
\begin{equation}
\vect{u}_j=
\frac{E_s(\phi(c_j))}
{\|E_s(\phi(c_j))\|_2},
\quad
\vect{v}_{j,i}=
\frac{E_s(z_{j,i})}
{\|E_s(z_{j,i})\|_2},
\quad i\in\{1,\ldots,N\},
\end{equation}
where \(\vect{u}_j\in\mathbb{R}^{d_t}\) is the normalized embedding of the templated description of category \(c_j\), and \(\vect{v}_{j,i}\in\mathbb{R}^{d_t}\) is the normalized embedding of the \(i\)-th candidate concept \(z_{j,i}\) of category \(c_j\). Since all vectors are L2-normalized, cosine similarity can be directly computed by inner products in the shared semantic embedding space. For each candidate concept \(z_{j,i}\), we compute its similarity to the target category and its maximum similarity to the other categories as
\begin{equation}
a_{j,i}=
\vect{v}_{j,i}^{\mathrm{T}}\vect{u}_j,
\quad
\eta_{j,i}=
\max_{\vect{u}\in\vect{U}_{-\!j}}
\vect{v}_{j,i}^{\mathrm{T}}\vect{u},
\end{equation}
where \(a_{j,i}\) denotes the similarity between \(z_{j,i}\) and the corresponding target category \(c_j\), \(\eta_{j,i}\) denotes its maximum similarity to other privacy-risk categories, and \(\vect{U}_{-\!j}\) contains the normalized embeddings of all privacy-risk categories except \(c_j\).

The discriminability score is then defined as
\begin{equation}
q_{j,i}=a_{j,i}-\eta_{j,i}.
\end{equation}

A larger \(q_{j,i}\) indicates a larger semantic similarity margin between the target category and the most similar non-target category. We rank the candidates of category \(c_j\) by \(q_{j,i}\) in descending order. The top \(M\) concepts are then retained to form the high-quality candidate set for category \(c_j\), defined as
\begin{equation}
Z_j^{\mathrm{top}}
=
\{z_{j,i}\mid i\in\mathcal{I}_j^{M}\},
\quad
\mathcal{I}_j^{M}
=
\operatorname{TopIdx}_{M}
\left(
\{q_{j,i}\}_{i=1}^{N}
\right).
\end{equation}

The top $M$ candidates are retained for diversity-based selection.
We set $M=\lceil 1.5K_c\rceil$, where $K_c$ denotes the final number of concepts retained per category.

\subsubsection{Clustering-based Selection}

The retained candidates are discriminative but may still be semantically redundant. Therefore, we apply K-Means to the normalized embeddings of \(Z_j^{\mathrm{top}}\).

The embeddings of the retained \(M\) candidate concepts are organized as
\begin{equation}
\vect{V}_j=
\begin{bmatrix}
\vect{v}_{j,i_1}^{\mathrm{T}}\\
\vect{v}_{j,i_2}^{\mathrm{T}}\\
\vdots\\
\vect{v}_{j,i_M}^{\mathrm{T}}
\end{bmatrix}
\in\mathbb{R}^{M\times d_t},
\quad
i_m\in\mathcal{I}_j^{M}.
\end{equation}

For each privacy-risk category, we set the number of K-Means clusters to \(K_c\) to partition the retained candidates into semantically distinct groups, yielding
\begin{equation}
\mathcal{G}_1,
\mathcal{G}_2,
\ldots,
\mathcal{G}_{K_c}
=
\mathrm{KMeans}(\vect{V}_j,K_c).
\end{equation}

For each cluster \(\mathcal{G}_r\), we select the concept nearest to its center \(\boldsymbol{\mu}_r\) as
\begin{equation}
z_{j,r}^{*}
=
\underset{z_{j,i}\in\mathcal{G}_r}{\arg\min}
\left\|
\vect{v}_{j,i}-\boldsymbol{\mu}_r
\right\|_2,
\quad
r\in\{1,\ldots,K_c\}.
\end{equation}

The selected cluster representatives form the final concept set for category \(c_j\), with one representative retained from each semantic cluster, given by
\begin{equation}
Z_j^*
=
\{z_{j,1}^*,z_{j,2}^*,\ldots,z_{j,K_c}^*\}.
\end{equation}

By applying this selection procedure to all privacy-risk categories, we obtain the privacy concept bank \(Z^*\) by concatenating the concept sets from all categories to construct a unified category-aware semantic concept space as
\begin{equation}
Z^*
=
\operatorname{Concat}
\left(
Z_1^*,Z_2^*,\ldots,Z_L^*
\right).
\end{equation}

The final privacy concept bank therefore contains a total of \(K=L K_c\) concept entries, with \(K_c\) concepts retained for each of the \(L\) privacy-risk categories for subsequent image--concept matching and prediction.

\subsection{Global-Region Concept Modeling}
\label{subsec:global_region_dual_visual_representation}

\subsubsection{Global and Regional Visual Features}

After the visual projection layer, the Long-CLIP visual encoder provides a class-token (CLS) representation and patch-token representations. The CLS representation captures image-level visual information, while the patch-token representations retain spatially localized features. We use this CLS representation as the global image feature and apply L2 normalization as
\begin{equation}
\bar{\vect{f}}_g
=
E_v^{\mathrm{CLS}}(I),
\qquad
\vect{f}_g
=
\frac{\bar{\vect{f}}_g}
{\|\bar{\vect{f}}_g\|_2},
\end{equation}
where \(\vect{f}_g\in\mathbb{R}^{d}\) denotes the L2-normalized global image representation.

We denote the projected and L2-normalized patch-token features as
\begin{equation}
\mathcal{H}
=
\{\vect{h}_t\}_{t=1}^{G_hG_w},
\end{equation}
where \(\vect{h}_t\in\mathbb{R}^{d}\), and \(G_h\) and \(G_w\) denote the numbers of patch tokens along the vertical and horizontal dimensions, respectively. The \(G_hG_w\) patch tokens are arranged according to their spatial positions to form a \(G_h\times G_w\) feature map.
\begin{equation}
\mathcal{H}_{\mathrm{map}}
\in
\mathbb{R}^{G_h\times G_w\times d}.
\end{equation}

To obtain regional representations, adaptive average pooling transforms the \(G_h\times G_w\) patch-token feature map into an \(R_h\times R_w\) regional grid, where \(R_h\) and \(R_w\) denote the numbers of regional bins along the vertical and horizontal dimensions, respectively, and \(R=R_hR_w\) denotes the total number of regions. Each regional bin aggregates patch-token features from its corresponding spatial area, providing localized representations for subsequent concept matching.

For the \(r\)-th region, \(r\in\{1,\ldots,R\}\), \(\Omega_r\) denotes the set of patch-token indices assigned to the corresponding pooling bin. The regional feature is computed as
\begin{equation}
\bar{\vect{f}}_{\mathrm{reg}}^r
=
\frac{1}{|\Omega_r|}
\sum_{t\in\Omega_r}\vect{h}_t,
\qquad
\vect{f}_{\mathrm{reg}}^r
=
\frac{\bar{\vect{f}}_{\mathrm{reg}}^r}
{\|\bar{\vect{f}}_{\mathrm{reg}}^r\|_2}.
\end{equation}

\subsubsection{Concept Responses and Softplus Mean Aggregation}

Before training, the selected concept set \(Z^*\) is encoded once by the frozen Long-CLIP text encoder to obtain the concept embedding matrix
\begin{equation}
\vect{T}
=
E_t(Z^{*})
=
[\vect{t}_1,\vect{t}_2,\ldots,\vect{t}_{K}]^\top
\in\mathbb{R}^{K\times d},
\end{equation}
where \(\vect{t}_n\in\mathbb{R}^{d}\) denotes the text embedding of the \(n\)-th concept,
\(n\in\{1,\ldots,K\}\). Since the concept embeddings are precomputed before training, \(\vect{T}\) remains fixed throughout training and subsequent privacy prediction.

The global response of the \(n\)-th concept is computed by cosine similarity as
\begin{equation}
m_{g,n}
=
\cos(\vect{f}_g,\vect{t}_n),
\qquad
\vect{m}_g
=
[m_{g,1},m_{g,2},\ldots,m_{g,K}].
\end{equation}

For the region branch, we compute the cosine similarity between each regional feature and each concept to measure region-specific concept responses as
\begin{equation}
m_{\mathrm{reg},n}^{r}
=
\cos(\vect{f}_{\mathrm{reg}}^r,\vect{t}_n),
\end{equation}
where \(m_{\mathrm{reg},n}^{r}\) denotes the cosine similarity between the \(r\)-th regional visual feature \(\vect{f}_{\mathrm{reg}}^r\) and the \(n\)-th concept embedding \(\vect{t}_n\), thereby quantifying how strongly the \(n\)-th concept responds within the \(r\)-th region for privacy recognition.

Since privacy cues may appear only in a few image regions, negative regional responses may cancel positive responses during aggregation. To reduce such cancellation, we apply a softplus transformation followed by mean pooling as
\begin{equation}
m_{\mathrm{reg},n}
=
\frac{1}{R}
\sum_{r=1}^{R}
\tau
\log
\left(
1+\exp
\left(
\frac{m_{\mathrm{reg},n}^{r}}{\tau}
\right)
\right),
\end{equation}
where \(\tau\) is the temperature parameter controlling the softplus transformation.

The aggregated regional concept responses are then collected into
\begin{equation}
\vect{m}_{\mathrm{reg}}
=
[m_{\mathrm{reg},1},m_{\mathrm{reg},2},\ldots,m_{\mathrm{reg},K}].
\end{equation}

Here, \(\vect{m}_g\) and \(\vect{m}_{\mathrm{reg}}\) denote global and regional concept responses.

\subsubsection{Concept-to-Class Mapping and Fusion}

Although generated category-wise, all selected privacy-related concepts are combined into a unified concept bank, whose responses are mapped to class scores using a shared learnable concept-to-class association matrix defined as
\begin{equation}
\vect{A}\in\mathbb{R}^{L\times K}.
\end{equation}
The association weights \(A_{j,n}\) characterize the direction and strength of concept--category associations, while the signed contribution of each concept to individual predictions also depends on its response and the scaling parameter.

The shared association matrix maps the global and regional concept responses to class scores for all privacy-risk categories simultaneously as
\begin{equation}
\vect{s}_g=\lambda \vect{m}_g\vect{A}^{\mathrm{T}},
\qquad
\vect{s}_{\mathrm{reg}}=\lambda \vect{m}_{\mathrm{reg}}\vect{A}^{\mathrm{T}},
\end{equation}
where \(\lambda\) is a learnable scaling parameter applied to the class scores, initialized to \(\ln(50)\), and jointly optimized with the other trainable parameters. Sharing \(\vect{A}\) ensures consistent concept-to-category semantics across both branches.

The final class logit is obtained by category-specific fusion as
\begin{equation}
s_j=(1-\alpha_j)s_{\mathrm{reg},j}+\alpha_js_{g,j}+b_j,
\end{equation}
where \(\alpha_j\in[0,1]\) controls the contribution of the global branch for category \(c_j\), and \(b_j\) is a learnable category bias initialized to the fixed constant \(-1.5\) and subsequently optimized during training. A larger \(\alpha_j\) indicates stronger reliance on global context, while a smaller value emphasizes regional evidence.

Each class logit is converted into a probability for multi-label prediction as
\begin{equation}
p_j=\sigma(s_j).
\end{equation}

The binary prediction for each privacy-risk category \(c_j\) is then defined as
\begin{equation}
\hat{y}_j=
\begin{cases}
1, & p_j>\gamma_j,\\
0, & p_j\leq\gamma_j,
\end{cases}
\end{equation}
where \(\gamma_j\) is the category-specific threshold for \(c_j\), selected on the validation set through category-wise calibration and fixed for test evaluation.

\subsection{Inference and Calibration}
\label{subsec:inference_protocol}

Fusion coefficients are periodically updated category-wise on the validation set during training, while the final decision thresholds are selected on the validation set before test evaluation. For each privacy-risk category \(c_j\), we evaluate a set of candidate fusion coefficients. Given a candidate coefficient \(\alpha\), the provisional bias-free probability for the current category is computed as
\begin{equation}
p_j^{(\alpha)}
=
\sigma\left(
(1-\alpha)s_{\mathrm{reg},j}
+
\alpha s_{g,j}
\right).
\end{equation}

For each candidate \(\alpha\), an inner decision threshold on the validation set is selected by maximizing the balanced accuracy (BA)--F1 score, defined as
\begin{equation}
S_{\mathrm{BA\text{-}F1}}
=
0.5\,\mathrm{BA}
+
0.5\,\mathrm{F1}.
\end{equation}

Candidate fusion coefficients are evaluated using the AP--F1 score, defined as
\begin{equation}
S_{\mathrm{AP\text{-}F1}}
=
0.5\,\mathrm{AP}
+
0.5\,\mathrm{F1},
\end{equation}
where AP is computed from the continuous prediction probabilities and F1 is evaluated using the corresponding optimal inner threshold. The fusion coefficient with the highest \(S_{\mathrm{AP\text{-}F1}}\) is selected as the category-specific coefficient \(\alpha_j\).

After determining \(\alpha_j\), the learned category bias \(b_j\) is added to the fused logit as defined above. The final decision threshold \(\gamma_j\) is then selected on the validation set by maximizing \(S_{\mathrm{BA\text{-}F1}}\). The selected \(\alpha_j\) and \(\gamma_j\) are fixed during test evaluation. The prediction for each privacy-risk category is given by
\begin{equation}
\hat{y}_j
=
\mathbf{1}
\left[
p_j>\gamma_j
\right].
\end{equation}

Safe/Public does not correspond to an independent concept-based classifier and is treated as a residual category. It is assigned only when none of the privacy-risk categories is activated during inference for each image, as expressed by
\begin{equation}
\hat{y}_{\mathrm{safe}}
=
1
\quad \mathrm{if} \quad
\forall c_j\in C_p,\;
p_j\leq\gamma_j.
\end{equation}
Thus, Safe/Public has neither an independent decision threshold nor a separate concept bank and is predicted only when all privacy-risk predictions are zero.

\subsection{Training Objective}
\label{subsec:training_objective}

During training, we optimize PriCBMs with Asymmetric Loss (ASL)~\cite{ridnik2021asymmetric}, which is designed for imbalanced multi-label classification by reducing the dominance of easy negative examples. For sample \(i\) and privacy-risk category \(j\), let \(y_{i,j}\in\{0,1\}\) denote the ground-truth label and \(p_{i,j}=\sigma(s_{i,j})\) denote the predicted probability. Following ASL, we define \(p_{i,j}^{+}=p_{i,j}\), \(p_{i,j}^{-}=1-p_{i,j}\), and apply probability clipping to negative labels as \(\bar{p}_{i,j}^{-}=\min(p_{i,j}^{-}+m,1)\).

The ASL objective is defined as
\begin{equation}
\begin{aligned}
\mathcal{L}_{\mathrm{ASL}}
&=
\frac{1}{B}\sum_{i=1}^{B}\sum_{j=1}^{L}
\Bigl[
-y_{i,j}(1-p_{i,j}^{+})^{\gamma_+}\log(p_{i,j}^{+})\\
&\hspace{2.6cm}
-(1-y_{i,j})(1-\bar{p}_{i,j}^{-})^{\gamma_-}
\log(\bar{p}_{i,j}^{-})
\Bigr].
\end{aligned}
\end{equation}

We further regularize the trainable Long-CLIP visual encoder by preserving its L2-normalized global image representation relative to that produced by a frozen reference visual encoder during optimization to reduce representation drift:
\begin{equation}
\mathcal{L}_{\mathrm{img}}
=
1-\frac{1}{B}\sum_{i=1}^{B}
\cos\!\left(
E_v(I_i),
E_v^{0}(I_i)
\right),
\end{equation}
where \(B\) is the batch size, while \(E_v(I_i)\) and \(E_v^{0}(I_i)\) denote the L2-normalized global image representations produced by the trainable and frozen Long-CLIP visual encoders, respectively, with their similarity measured using cosine similarity.

The overall training objective is
\begin{equation}
\mathcal{L}
=
\mathcal{L}_{\mathrm{ASL}}
+
\beta_I\mathcal{L}_{\mathrm{img}}.
\end{equation}

The Long-CLIP text encoder remains frozen throughout training. Safe/Public samples are used as negative samples for all privacy-risk categories, and no independent Safe/Public classifier or Safe/Public concept loss is introduced.

\section{Experiments}
\label{sec:experiments}

\subsection{Datasets and Label Construction}
\label{subsec:datasets_label_construction}

We evaluate PriCBMs on two multi-label visual privacy recognition datasets, VISPR and FineViP~\cite{ref30}. VISPR contains 68 attributes, including 67 privacy-risk attributes and one Safe attribute, which is assigned when no privacy-risk attribute is present. We use a fixed 7:2:1 training/validation/test split for all VISPR experiments, including reproduced baselines and ablation studies, to ensure consistent comparisons. VISPR is released under the Creative Commons Attribution-NonCommercial 4.0 International (CC BY-NC 4.0) license, while individual images retain their original licenses.

FineViP provides 22 High- and Low-risk labels corresponding to 11 privacy categories. We follow the official splits and retain retrieved images from provided URLs. Since we focus on identifying privacy categories rather than distinguishing risk levels, we merge the High- and Low-risk labels of each category as
\begin{equation}
y_c = y_c^{\mathrm{high}} \vee y_c^{\mathrm{low}},
\end{equation}
where \(y_c^{\mathrm{high}}\) and \(y_c^{\mathrm{low}}\) denote the original high- and low-risk labels, respectively. The resulting label space contains 11 privacy-risk categories and one Public category, treated as Safe/Public in our formulation. The FineViP annotations, taxonomy, and related resources are released under the Creative Commons Attribution-NonCommercial-ShareAlike 4.0 International (CC BY-NC-SA 4.0) license, while individual images remain subject to their original licenses and terms.

\subsection{Baselines and Comparison Groups}
\label{subsec:baselines_comparison_groups}

We organize the comparison experiments into four groups. First, we compare PriCBMs with representative multi-label visual privacy recognition methods, including PrivAttNetMLC~\cite{chen2021privattnet} and PSSA~\cite{jiang2025connecting}. PrivAttNetMLC employs an attention-based architecture to capture privacy-relevant visual regions and predict multiple privacy attributes within an image. PSSA introduces privacy-specific spatial attention to extract category-dependent visual features, allowing different privacy categories to focus on distinct regions of the same image.

Second, we compare PriCBMs with representative general multi-label classification methods, including Two-Way Multi-Label Loss (TWL)~\cite{kobayashi2023two}, MMBipolar~\cite{mishra2023memory}, and Bipolar-based methods~\cite{ref25}, including BipolarBCE, BipolarFocal, BipolarTWL, and BipolarSM. We also include BCE and Focal Loss~\cite{lin2017focal} as standard loss-based baselines. Following the experimental settings of TWL~\cite{kobayashi2023two} and BipolarFocal~\cite{ref25}, we adopt the same four visual backbones, namely ResNet50 and ResNet101~\cite{ref15}, ResNeXt~\cite{ref16}, and RegNet-Y-32gf~\cite{ref17}, to evaluate BCE, Focal, TWL, MMBipolar, and the Bipolar-based methods, thereby maintaining a consistent backbone setting across these baselines.

Third, we compare PriCBMs with representative multimodal large language models (MLLMs), which are grouped according to how they are accessed during evaluation. The locally deployed models include Qwen2.5-VL~\cite{qwen25vl2025}, Qwen3.5~\cite{qwen32025}, and DeepSeek-VL2~\cite{deepseekvl22024}, while the API-accessed models include GPT-5.5~\cite{OpenAI_gpt55}, Gemini-3.6-Flash~\cite{google_gemini36flash}, and Kimi K3~\cite{kimi2026k3}.

Fourth, since binary private/public classification remains common in visual privacy recognition, we additionally evaluate a task-specific binary variant, termed PriCBMs-Binary, for broader comparison. PriCBMs-Binary is trained independently under the Safe/Public-versus-Privacy label space rather than obtained by directly collapsing the original multi-label predictions.

Since the multi-label baselines use different visual backbones and pretraining schemes, the comparison is intended to reflect representative performance under commonly used settings rather than a strictly backbone-controlled comparison. To isolate the effect of the proposed formulation, we further provide controlled comparisons under the same Long-CLIP backbone in Section~\ref{subsubsec:model_formulation_summary}.

\subsection{Implementation and Training Details}
\label{subsec:implementation_training_details}

We adopt Long-CLIP with ViT-L/14 as the unified vision-language encoder and initialize it from the pretrained longclip-L model. Images are resized to \(224\times224\) using the default CLIP preprocessing. The text encoder remains frozen throughout training, while the visual encoder is trainable.

For the regional branch, adaptive average pooling produces a \(3\times3\) regional grid, resulting in \(R=9\) regions. Regional concept responses are aggregated using the softplus mean with a temperature of \(\tau=0.07\). For each privacy-risk category, we remove duplicate and invalid outputs and retain \(N = 100\) valid candidate concepts. From this candidate pool, we select \(K_c = 30\) concepts through semantic discriminability and diversity selection for subsequent modeling.

We train PriCBMs using AdamW for 100 epochs. The learning rate is set to \(2\times10^{-5}\) for the visual encoder and \(5\times10^{-5}\) for the remaining trainable parameters, with an effective batch size of 32. We use Asymmetric Loss for multi-label classification and set the image feature regularization weight to \(\beta_I=0.01\). 

All model configurations are selected using the validation set and fixed before test evaluation. Specifically, the validation set is used to determine the category-specific fusion coefficients and decision thresholds, as well as the number of retained concepts per category and the number of image regions. The test set is used only for performance reporting and is not involved in hyperparameter selection. Test-set sensitivity analyses for the number of retained concepts and the number of regions are provided in appendix~\ref{appendix:configuration_analysis}. These analyses are reported only for post-hoc robustness evaluation and are not used for model selection.

\subsection{Evaluation Metrics}
\label{subsec:evaluation_metrics}

Following the standard evaluation protocol for multi-label learning~\cite{zhang2013review}, we report MAP-class and MAP-sample to evaluate ranking performance, together with Micro-F1 and Macro-F1 to evaluate classification performance.

Specifically, MAP-class computes the average precision for each category and then averages over categories, while MAP-sample computes the average precision for each image and then averages over samples. For all evaluated methods, MAP-class and MAP-sample are computed over the complete label space, including the residual Safe/Public category. For ranking-based evaluation, Safe/Public is assigned as the ground-truth label when no privacy-risk category is present, and its ranking score, derived from privacy-category probabilities, is defined as
\begin{equation}
p_{\mathrm{safe}}
=
1-\max_{c_j\in C_p} p_j.
\label{eq:safe_confidence}
\end{equation}

After the decision thresholds are applied, Micro-F1 and Macro-F1 are also computed over the complete label space, including the Safe/Public category. Micro-F1 aggregates true positives, false positives, and false negatives over all labels, while Macro-F1 averages the F1 scores across labels and therefore better reflects class-balanced performance. Unless otherwise specified, all metric values reported in the tables are presented in percentage form to facilitate comparison.

For the additional binary classification experiments with PriCBMs-Binary, we report accuracy, balanced accuracy, and Macro-F1 as overall evaluation metrics. We also report class-wise precision, recall, and F1-score for the Public and Privacy classes to provide a more detailed assessment under class imbalance.

\subsection{Results}
\label{subsec:results}

\subsubsection{Multi-label Classification Comparison}

As shown in Tables~\ref{tab:comparison_vispr} and~\ref{tab:comparison_finevip}, PriCBMs achieve the best performance across all four metrics on both VISPR and FineViP. The gains are particularly pronounced in MAP-class and Macro-F1. On VISPR, PriCBMs reach 67.28\% MAP-class and 63.51\% Macro-F1, outperforming the strongest baselines by 11.72 and 22.43 percentage points, respectively. On FineViP, the corresponding improvements are 10.46 and 12.39 percentage points. These consistent gains across both datasets indicate improved category-level prediction, as reflected in MAP-class and Macro-F1. The higher Macro-F1 indicates improved average performance across all categories, including Safe/Public, under equal category weighting.

PriCBMs also obtain the best MAP-sample and Micro-F1 results. The improvements are 1.34 and 2.02 percentage points on VISPR, compared with 5.70 and 10.33 percentage points on FineViP. This difference may be related to the larger label space of VISPR, which contains 68 categories compared with 12 in FineViP. A larger label space may increase the difficulty of sample-level ranking and joint multi-label prediction, which may explain why the performance gains on VISPR are more pronounced in category-level metrics.

PriCBMs further outperform the privacy-oriented PrivAttNetMLC and PSSA across all four metrics. Unlike these attention-based methods, PriCBMs explicitly model global and regional privacy concept responses and combine them through category-specific fusion. This allows complementary evidence from different spatial levels to be represented explicitly in the concept space. The results show that PriCBMs achieve better multi-label classification performance than the evaluated privacy-oriented baselines under the reported settings.

\begin{table}[t]
\caption{Comparison results on the VISPR dataset.}
\label{tab:comparison_vispr}
\begin{threeparttable}
\centering
\rmfamily
{\fontsize{10pt}{12pt}\selectfont
\setlength{\tabcolsep}{4pt}
\renewcommand{\arraystretch}{1.02}
\setlength{\aboverulesep}{0.35ex}
\setlength{\belowrulesep}{0.35ex}
\setlength{\cmidrulesep}{0.20ex}

\begin{tabular*}{\linewidth}{@{\extracolsep{\fill}} llcccc @{}}
\toprule
\textbf{Backbone} &
\textbf{Method} &
\textbf{MAP-c} &
\textbf{MAP-s} &
\textbf{Mic-F1} &
\textbf{Mac-F1} \\
\midrule

\multirow{8}{*}{ResNet50}
& BCE          & 49.54 & 85.36 & 78.20 & 34.80 \\
& Focal        & 48.84 & 84.71 & 77.39 & 35.96 \\
& TWL          & 50.02 & 81.08 & 66.34 & 38.24 \\
& MMBipolar    & 53.51 & 84.39 & 78.96 & 36.13 \\
& BipolarBCE   & 51.39 & 84.86 & 78.59 & 38.91 \\
& BipolarFocal & 50.45 & 83.86 & 77.77 & 39.90 \\
& BipolarTWL   & 47.81 & 84.16 & 32.44 & 18.84 \\
& BipolarSM    & 50.32 & 84.14 & 78.34 & 31.78 \\
\midrule

\multirow{8}{*}{ResNet101}
& BCE          & 50.03 & 85.31 & 77.91 & 35.53 \\
& Focal        & 49.32 & 84.49 & 77.92 & 36.54 \\
& TWL          & 49.42 & 80.16 & 65.81 & 38.05 \\
& MMBipolar    & 52.12 & 83.60 & 78.91 & 39.80 \\
& BipolarBCE   & 51.19 & 84.95 & 78.77 & 39.64 \\
& BipolarFocal & 50.18 & 84.36 & 77.97 & 41.08 \\
& BipolarTWL   & 48.74 & 83.00 & 32.84 & 19.08 \\
& BipolarSM    & 49.04 & 83.97 & 78.31 & 32.18 \\
\midrule

\multirow{8}{*}{ResNeXt}
& BCE          & 50.08 & 85.39 & 78.30 & 34.98 \\
& Focal        & 48.25 & 84.55 & 77.68 & 34.68 \\
& TWL          & 50.84 & 81.68 & 68.61 & 39.62 \\
& MMBipolar    & 53.98 & 85.30 & 78.57 & 37.19 \\
& BipolarBCE   & 51.88 & 85.25 & 78.37 & 37.56 \\
& BipolarFocal & 50.77 & 85.23 & 77.44 & 38.70 \\
& BipolarTWL   & 49.02 & 84.49 & 33.20 & 18.87 \\
& BipolarSM    & 50.24 & 83.92 & 78.42 & 33.18 \\
\midrule

\multirow{8}{*}{RegNet-Y-32gf}
& BCE          & 49.17 & 85.87 & 79.20 & 34.27 \\
& Focal        & 49.83 & 85.88 & 78.17 & 34.45 \\
& TWL          & 55.56 & 82.62 & 51.19 & 29.46 \\
& MMBipolar    & 54.69 & 85.88 & 80.09 & 39.55 \\
& BipolarBCE   & 51.30 & 85.96 & 79.50 & 38.24 \\
& BipolarFocal & 50.77 & 85.67 & 78.86 & 39.64 \\
& BipolarTWL   & 52.58 & 85.04 & 33.24 & 19.61 \\
& BipolarSM    & 50.40 & 84.43 & 79.02 & 33.42 \\
\midrule

\multicolumn{2}{l}{\hspace{1.6cm}PSSA}
& 33.91 & 78.82 & 71.80 & 23.42 \\
\multicolumn{2}{l}{\hspace{1.6cm}PrivAttNetMLC}
& 39.69 & 84.90 & 74.85 & 25.83 \\
\midrule

\multicolumn{2}{l}{\hspace{1.6cm}\textbf{PriCBMs}}
& \textbf{67.28}
& \textbf{87.30}
& \textbf{82.11}
& \textbf{63.51} \\
\bottomrule
\end{tabular*}
}

\vspace{-0.3mm}

\begin{tablenotes}[flushleft]
\footnotesize
\rmfamily
\item MAP-c/s: MAP-class/sample; Mic-F1/Mac-F1: Micro-/Macro-F1. Best results are in bold.
\end{tablenotes}

\end{threeparttable}
\end{table}

\begin{table}[t]
\caption{Comparison results on the FineViP dataset.}
\label{tab:comparison_finevip}
\begin{threeparttable}
\centering
\rmfamily
{\fontsize{10pt}{12pt}\selectfont
\setlength{\tabcolsep}{4pt}
\renewcommand{\arraystretch}{1.02}
\setlength{\aboverulesep}{0.35ex}
\setlength{\belowrulesep}{0.35ex}
\setlength{\cmidrulesep}{0.20ex}

\begin{tabular*}{\linewidth}{@{\extracolsep{\fill}} llcccc @{}}
\toprule
\textbf{Backbone} &
\textbf{Method} &
\textbf{MAP-c} &
\textbf{MAP-s} &
\textbf{Mic-F1} &
\textbf{Mac-F1} \\
\midrule

\multirow{8}{*}{ResNet50}
& BCE          & 67.46 & 80.69 & 68.40 & 62.03 \\
& Focal        & 66.90 & 80.26 & 67.66 & 61.71 \\
& TWL          & 66.96 & 81.08 & 60.41 & 57.29 \\
& MMBipolar    & 65.11 & 79.22 & 66.54 & 59.98 \\
& BipolarBCE   & 67.38 & 81.07 & 67.91 & 62.34 \\
& BipolarFocal & 67.70 & 81.20 & 68.11 & 62.01 \\
& BipolarTWL   & 67.09 & 81.21 & 60.31 & 57.71 \\
& BipolarSM    & 66.17 & 80.50 & 68.31 & 62.29 \\
\midrule

\multirow{8}{*}{ResNet101}
& BCE          & 68.50 & 81.23 & 68.80 & 62.86 \\
& Focal        & 68.28 & 80.65 & 68.25 & 62.35 \\
& TWL          & 68.46 & 82.15 & 60.20 & 57.73 \\
& MMBipolar    & 62.29 & 79.12 & 67.94 & 62.10 \\
& BipolarBCE   & 69.45 & 81.51 & 68.76 & 63.92 \\
& BipolarFocal & 68.48 & 81.12 & 68.14 & 62.15 \\
& BipolarTWL   & 68.19 & 81.37 & 59.15 & 56.65 \\
& BipolarSM    & 67.14 & 81.01 & 68.75 & 62.75 \\
\midrule

\multirow{8}{*}{ResNeXt}
& BCE          & 67.79 & 80.38 & 67.31 & 62.25 \\
& Focal        & 67.63 & 80.58 & 67.87 & 62.47 \\
& TWL          & 67.39 & 80.79 & 58.05 & 55.36 \\
& MMBipolar    & 66.69 & 80.09 & 68.16 & 61.88 \\
& BipolarBCE   & 68.55 & 80.98 & 68.51 & 63.00 \\
& BipolarFocal & 69.30 & 81.35 & 67.85 & 62.53 \\
& BipolarTWL   & 67.15 & 81.10 & 57.65 & 55.03 \\
& BipolarSM    & 67.17 & 80.87 & 68.97 & 62.83 \\
\midrule

\multirow{8}{*}{RegNet-Y-32gf}
& BCE          & 71.87 & 83.24 & 71.40 & 67.03 \\
& Focal        & 72.25 & 82.86 & 71.05 & 66.30 \\
& TWL          & 71.47 & 83.51 & 60.95 & 58.60 \\
& MMBipolar    & 67.64 & 82.23 & 71.69 & 66.96 \\
& BipolarBCE   & 72.00 & 83.01 & 71.57 & 66.86 \\
& BipolarFocal & 71.70 & 82.72 & 70.64 & 65.62 \\
& BipolarTWL   & 71.83 & 83.06 & 61.99 & 59.67 \\
& BipolarSM    & 69.78 & 83.10 & 72.00 & 66.94 \\
\midrule

\multicolumn{2}{l}{\hspace{1.6cm}PSSA}
& 41.85 & 65.67 & 37.26 & 28.86 \\
\multicolumn{2}{l}{\hspace{1.6cm}PrivAttNetMLC}
& 62.89 & 76.48 & 59.12 & 53.84 \\
\midrule

\multicolumn{2}{l}{\hspace{1.6cm}\textbf{PriCBMs}}
& \textbf{82.71}
& \textbf{89.21}
& \textbf{82.33}
& \textbf{79.42} \\
\bottomrule
\end{tabular*}
}

\vspace{-0.3mm}

\begin{tablenotes}[flushleft]
\footnotesize
\rmfamily
\item MAP-c/s: MAP-class/sample; Mic-F1/Mac-F1: Micro-/Macro-F1. Best results are in bold.
\end{tablenotes}

\end{threeparttable}
\end{table}

\FloatBarrier

\subsubsection{Comparison with Multimodal Large Language Models}

We compare PriCBMs with multimodal large language models (MLLMs) to examine whether they can directly perform multi-label visual privacy recognition. Each MLLM is prompted to identify all privacy-related categories present in the image from the predefined label space and return the corresponding multi-label predictions. If a model refuses to answer, the sample is counted as a refusal case; otherwise, its returned labels are evaluated. Because refusal reduces prediction coverage and affects the subset on which classification performance is evaluated, we report refusal rates and conditional classification performance computed on non-refused samples as separate outcomes for clearer interpretation.

\begin{center}
\begin{minipage}{0.9\linewidth}
\centering
\captionof{table}{Comparison with open-source multimodal large language models.}
\label{tab:open_mllm_comparison}
\begin{threeparttable}
\centering
{\normalsize
\setlength{\tabcolsep}{2.0pt}
\renewcommand{\arraystretch}{1.0}
\setlength{\aboverulesep}{0.25ex}
\setlength{\belowrulesep}{0.25ex}
\setlength{\cmidrulesep}{0.15ex}
\begin{tabular*}{\linewidth}{@{\extracolsep{\fill}} l l c c @{}}
\toprule
\textbf{Data} &
\textbf{Model} &
\textbf{Mic-F1} &
\textbf{Mac-F1} \\
\midrule
\multirow{5}{*}{VISPR}
& Q2.5-32B & 43.20 & 40.64 \\
& Q2.5-72B & 39.11 & 40.71 \\
& Q3.5-27B & 41.75 & 22.07 \\
& DSVL2    & 17.51 &  6.71 \\
& \textbf{PriCBMs} & \textbf{82.11} & \textbf{63.51} \\
\midrule
\multirow{5}{*}{FineViP}
& Q2.5-32B & 41.84 & 39.23 \\
& Q2.5-72B & 42.55 & 42.53 \\
& Q3.5-27B & 44.39 & 45.19 \\
& DSVL2    & 19.03 & 18.51 \\
& \textbf{PriCBMs} & \textbf{82.33} & \textbf{79.42} \\
\bottomrule
\end{tabular*}
}
\vspace{0.5mm}
\begin{tablenotes}[flushleft]
\footnotesize
\item \parbox{\dimexpr\linewidth-3pt\relax}{Q2.5-32B, Q2.5-72B, and Q3.5-27B denote Qwen2.5-VL-32B-Instruct, Qwen2.5-VL-72B-Instruct, and Qwen3.5-27B, respectively. DSVL2 denotes DeepSeek-VL2.}
\end{tablenotes}
\end{threeparttable}
\end{minipage}
\end{center}
\FloatBarrier

\begin{center}
\begin{minipage}{0.9\linewidth}
\centering
\captionof{table}{Comparison with API-accessed multimodal large language models.}
\label{tab:closed_mllm_comparison}

\begin{threeparttable}
\centering
{\normalsize
\setlength{\tabcolsep}{2.0pt}
\renewcommand{\arraystretch}{1.0}
\setlength{\aboverulesep}{0.25ex}
\setlength{\belowrulesep}{0.25ex}
\setlength{\cmidrulesep}{0.15ex}

\begin{tabular*}{\linewidth}{@{\extracolsep{\fill}} l l c c c @{}}
\toprule
\textbf{Data} &
\textbf{Model} &
\textbf{Mic-F1} &
\textbf{Mac-F1} &
\textbf{Refusal (\%)} \\
\midrule
\multirow{4}{*}{VISPR}
& GPT-5.5 & 82.71 & 10.38 & 74.15 \\
& Gemini-3.6-Flash & -- & -- & 92.20 \\
& Kimi K3 & 85.20 & 8.54 & 80.56 \\
& PriCBMs & 82.11 & 63.51 & \textbf{0.00} \\
\midrule
\multirow{4}{*}{FineViP}
& GPT-5.5 & 62.27 & 49.13 & 65.41 \\
& Gemini-3.6-Flash & -- & -- & 99.21 \\
& Kimi K3 & -- & -- & 96.42 \\
& PriCBMs & 82.33 & 79.42 & \textbf{0.00} \\
\bottomrule
\end{tabular*}
}

\vspace{0.5mm}
\begin{tablenotes}[flushleft]
\footnotesize
\item \parbox{\dimexpr\linewidth-3pt\relax}{``--'' indicates that classification metrics are omitted when the refusal rate exceeds 90\%. For API-accessed MLLMs, Micro-F1 and Macro-F1 are computed only on the non-refused subset. For Refusal, lower is better.}
\end{tablenotes}
\end{threeparttable}
\end{minipage}
\end{center}

\FloatBarrier

For open-source MLLMs, the comparison results are shown in Table~\ref{tab:open_mllm_comparison}.
These models provide predictions for the full test set and are therefore
evaluated under the full-coverage setting for direct comparison with the
proposed method.

As shown in Table~\ref{tab:open_mllm_comparison}, PriCBMs consistently outperform the evaluated open-source general-purpose MLLMs on both VISPR and FineViP under the full-coverage setting. On VISPR, PriCBMs achieve 82.11\% Micro-F1 and 63.51\% Macro-F1, while the strongest open-source MLLMs achieve 43.20\% Micro-F1 and 40.71\% Macro-F1, respectively. On FineViP, PriCBMs achieve 82.33\% Micro-F1 and 79.42\% Macro-F1, substantially outperforming the evaluated open-source MLLMs on both metrics. These results indicate that, under the evaluated prompting setup, the tested open-source MLLMs are less effective than the proposed task-specific concept-based framework on VISPR and FineViP.

For API-accessed MLLMs, the comparison results are shown in Table~\ref{tab:closed_mllm_comparison}. Unlike the locally deployed models, which provide predictions for all images in our evaluation, API-accessed MLLMs may refuse to answer privacy-related queries, potentially due to built-in safety mechanisms.

As shown in Table~\ref{tab:closed_mllm_comparison}, API-accessed MLLMs exhibit high refusal rates. GPT-5.5 refuses 74.15\% of VISPR samples and 65.41\% of FineViP samples, while Kimi K3 refuses 80.56\% and 96.42\% of samples on the two datasets, respectively. Gemini-3.6-Flash produces even fewer valid predictions, with refusal rates of 92.20\% on VISPR and 99.21\% on FineViP. These results show that API-accessed MLLMs have limited prediction coverage in visual privacy recognition.

The Micro-F1 and Macro-F1 scores of API-accessed MLLMs should therefore be interpreted as conditional performance on the non-refused subset rather than full-coverage performance. These scores are computed only on samples for which valid predictions are returned and are therefore not directly comparable with the full-coverage performance of PriCBMs under the same evaluation setting. The current analysis does not establish how refusal changes the class composition of the evaluated subset. Therefore, high conditional performance on valid predictions does not necessarily indicate reliable recognition across the full test set.

\subsubsection{Effect of Concept Bottleneck and Global--Region Modeling}
\label{subsubsec:model_formulation_summary}

To examine how the models relate to the two challenges, we compare four model formulations on VISPR. L-CLIP-Linear and L-CLIP-MLP directly predict privacy labels from global Long-CLIP visual features using a linear classifier and a multi-layer perceptron, respectively, without explicit concept-based intermediate representations. In contrast, L-CLIP-CBMs introduce privacy-related concepts between visual features and privacy labels, establishing an explicit image--concept--category process while relying solely on global image features. PriCBMs further extend this formulation by incorporating region-level concept responses and category-specific global-region fusion to model complementary privacy evidence. For a fair comparison, all four formulations use the same VISPR split, Long-CLIP backbone, training protocol, and validation-based threshold calibration procedure, with Safe/Public treated as a residual category during inference. The results of this controlled comparison across all four metrics are reported in Table~\ref{tab:model_formulation_summary}.

The comparison between direct image-to-label classifiers and L-CLIP-CBMs examines the effect of introducing privacy-related concepts as intermediate representations. As shown in Table~\ref{tab:model_formulation_summary}, L-CLIP-CBMs achieve 67.05\% MAP-class and 56.80\% Macro-F1, surpassing L-CLIP-Linear by 0.55 and 12.29 percentage points, and L-CLIP-MLP by 18.05 and 25.77 percentage points, respectively. These results suggest that replacing direct image-to-label prediction with a concept-mediated process, in which the model first identifies category-relevant privacy cues through visual--concept matching and then predicts privacy categories from the resulting concept responses, can improve classification performance. This finding is consistent with our research motivation.

\begin{center}
\begin{minipage}{1.0\linewidth}
\centering
\captionof{table}{Comparison of different model formulations on VISPR.}
\label{tab:model_formulation_summary}

{\normalsize
\setlength{\tabcolsep}{4pt}
\renewcommand{\arraystretch}{1.0}
\setlength{\aboverulesep}{0.25ex}
\setlength{\belowrulesep}{0.25ex}
\setlength{\cmidrulesep}{0.15ex}

\begin{tabular*}{0.95\linewidth}{@{\extracolsep{\fill}} lcccc @{}}
\toprule
\textbf{Model} &
\textbf{MAP-c} &
\textbf{MAP-s} &
\textbf{Mic-F1} &
\textbf{Mac-F1} \\
\midrule
L-CLIP-Linear & 66.50 & 86.89 & 65.61 & 44.51 \\
L-CLIP-MLP    & 49.00 & 86.07 & 43.00 & 31.03 \\
L-CLIP-CBMs   & 67.05 & 87.05 & 77.00 & 56.80 \\
\textbf{PriCBMs} &
\textbf{67.28} & \textbf{87.30} & \textbf{82.11} & \textbf{63.51} \\
\bottomrule
\end{tabular*}
}

\vspace{0.5mm}

\begin{minipage}{0.95\linewidth}
\footnotesize
\noindent
L-CLIP denotes Long-CLIP. L-CLIP-CBMs denote the global-only concept bottleneck baseline without global-region modeling, whereas PriCBMs denote the full global-region model. All four formulations use the same Safe/Public residual-category treatment during inference.
\end{minipage}

\end{minipage}
\end{center}

\FloatBarrier

The comparison between L-CLIP-CBMs and PriCBMs further examines the benefit of incorporating region-level privacy evidence beyond global concept representations. PriCBMs increase Micro-F1 from 77.00\% to 82.11\% and Macro-F1 from 56.80\% to 63.51\%, while also improving MAP-class from 67.05\% to 67.28\% and MAP-sample from 87.05\% to 87.30\%. These improvements suggest that integrating global- and region-level concept responses provides complementary information for multi-label privacy prediction. The results are consistent with our motivation to complement global representations with localized privacy evidence.

\FloatBarrier

\subsubsection{Effect of the Vision-Language Encoder}

We evaluate the effect of vision-language encoders by using Long-CLIP, CLIP, and LLM2CLIP~\cite{llm2clip2026} in PriCBMs while keeping other model components unchanged. This comparison focuses on differences in overall multi-label privacy recognition performance among the three encoder configurations within the same PriCBMs framework. The results are shown in Fig.~\ref{fig:encoder_ablation}.

Long-CLIP achieves the best performance across all four metrics on VISPR, with 67.28\% MAP-class, 87.30\% MAP-sample, 82.11\% Micro-F1, and 63.51\% Macro-F1. Compared with the second-best result on each metric, Long-CLIP improves these metrics by 7.27, 2.28, 3.92, and 3.21 percentage points, respectively.

\begin{center}
\begin{minipage}{1.0\linewidth}
\centering
\includegraphics[width=0.9\linewidth]{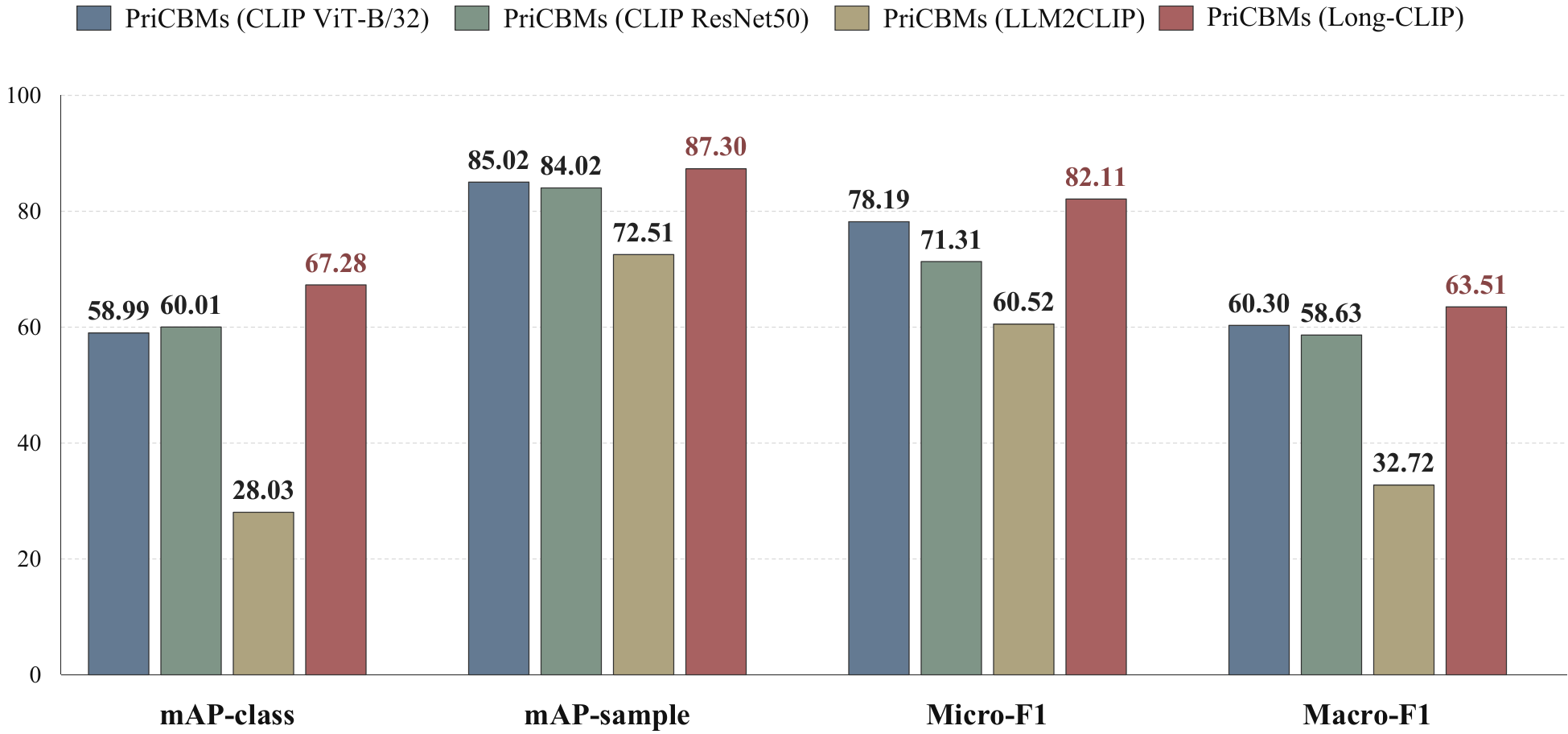}
\captionof{figure}{Performance comparison of different vision-language encoders on VISPR.}
\label{fig:encoder_ablation}
\end{minipage}
\end{center}

\FloatBarrier

\subsubsection{Effect of Concept Selection}

Concept selection builds a compact and interpretable semantic bottleneck for privacy prediction. To assess its two stages, we compare PriCBMs with three ablated variants, each starting from the same 100 candidate concepts and retaining 30 per category. The full method selects the top 45 by semantic discriminability and uses K-Means to retain 30. ``w/o Semantic Discriminability Selection'' applies K-Means directly to all 100 candidates; ``w/o Clustering-based Selection'' retains the top 30 by discriminability; and ``w/o Concept Selection'' randomly selects 30. The results are reported in Fig.~\ref{fig:concept_selection_ablation} for a direct comparison.

As shown in Fig.~\ref{fig:concept_selection_ablation}, our selection method achieves the best results on three out of four metrics and remains competitive on MAP-class. A further concept-level quality comparison is provided for additional analysis in Section~\ref{sec:interpretability_comparison}.

\begin{center}
\begin{minipage}{1.0\linewidth}
\centering
\setstretch{1}
\includegraphics[
width=0.95\linewidth,
keepaspectratio
]{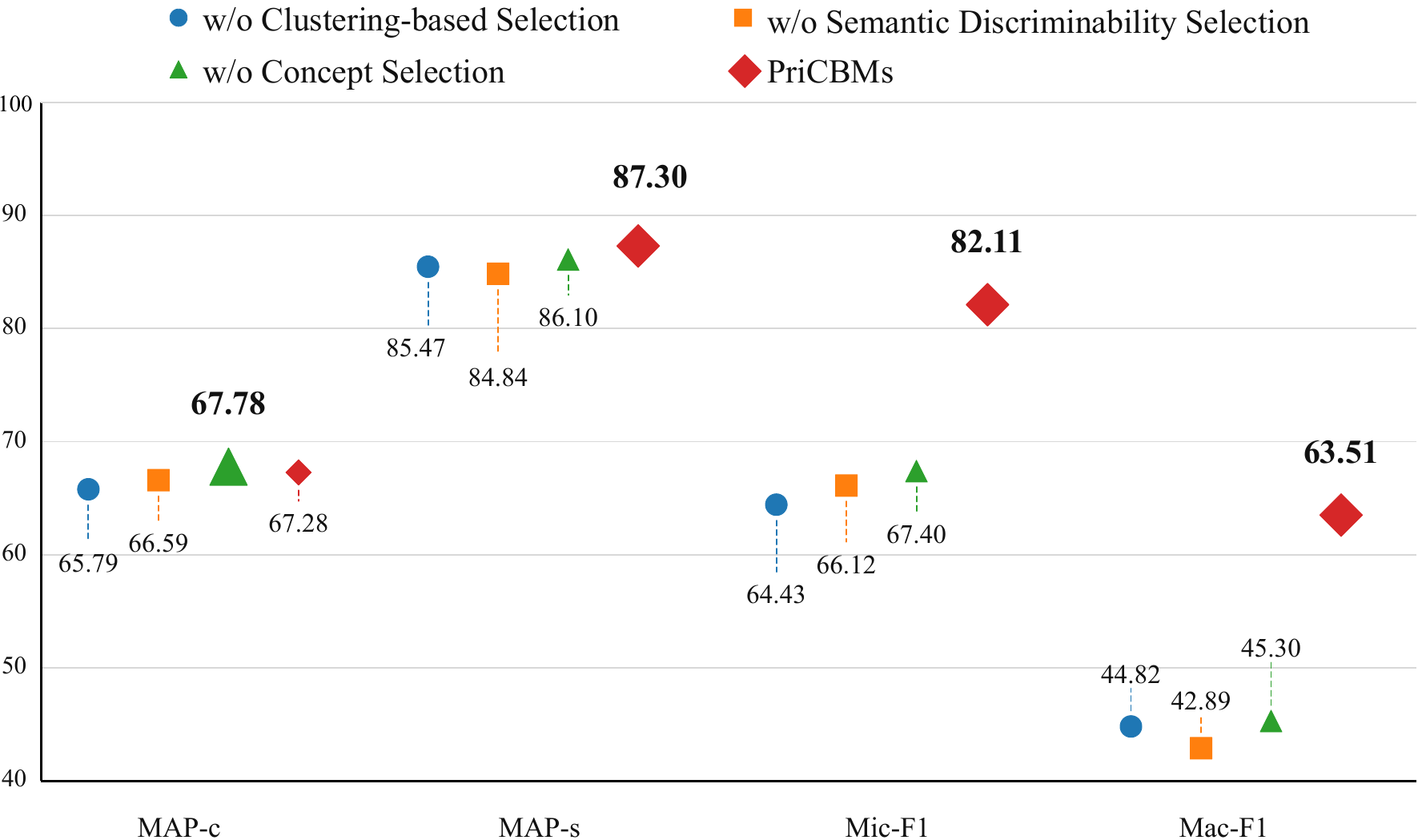}
\vspace{0.2em}
\captionof{figure}{Ablation results of the concept selection strategy on VISPR.}
\label{fig:concept_selection_ablation}
\end{minipage}
\end{center}

\FloatBarrier

\section{Visualization and Interpretability Analysis}
\label{sec:visualization}

\subsection{Visualization of Selected Concepts}
\label{subsec:selected_concepts}

To qualitatively assess the selected concepts, we select 12 images from VISPR~\cite{ref3} covering privacy-related scenarios involving textual and document information, facial and appearance cues, interpersonal interactions, contextual and group activities, and sensitive visual content. We visualize image-concept associations through the top three concepts ranked by combined global and regional responses. Due to space limitations, Fig.~\ref{fig:xai_concept} shows the first four examples, with the full set provided in appendix~\ref{appendix:selected_concepts_full}. For the sampled images, the selected concepts are generally consistent with the visible image content and privacy-related semantics, supporting their qualitative semantic relevance on these examples.

\begin{center}
\setstretch{1}
\includegraphics[
width=\linewidth,
height=0.60\textheight,
keepaspectratio
]{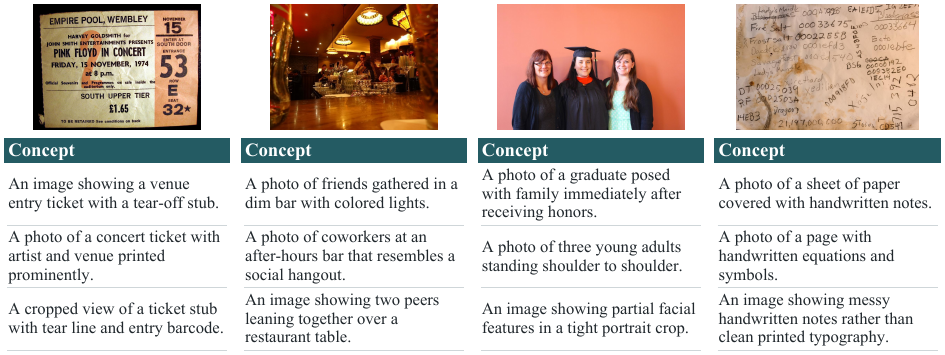}
\vspace{0.2em}
\captionof{figure}{Visualization of selected concepts for the first four representative VISPR images~\cite{ref3}. The top three are ranked by combined global and regional responses.}
\label{fig:xai_concept}
\end{center}

\FloatBarrier

\subsection{Quantitative Concept Quality Evaluation}
\label{sec:interpretability_comparison}

In addition to qualitative visualization, we conduct a quantitative concept-level evaluation to compare the semantic quality of the generated concepts on the sampled examples. We use the same sampled images considered in Section~\ref{subsec:selected_concepts}. For each sampled image, the top three concepts from each method are evaluated. The five compared methods are LaBo~\cite{ref13}, w/o Semantic Discriminability Selection, w/o Clustering-based Selection, w/o Concept Selection, and PriCBMs.

Each concept is rated on a five-point scale, where higher scores indicate better semantic correctness, visual grounding, and explanatory specificity. The scoring rubric is provided in Appendix~\ref{appendix:concept_level_scores}. To reduce reliance on a single evaluator, we use three vision-language model evaluators, GPT-5.5, Gemini-3.6-Flash, and Claude-Opus-5~\cite{anthropic_claudeopus5}. For a robust assessment, the Concept Quality Score (CQS) is defined as the average score across all evaluated concepts and evaluators, given by
\begin{equation}
\mathrm{CQS}
=
\frac{1}{|\mathcal{C}|E}
\sum_{c\in\mathcal{C}}
\sum_{e=1}^{E}
q_e(c),
\end{equation}
where \(E=3\), \(\mathcal{C}\) denotes evaluated image--concept pairs per method across the twelve sampled images, and \(q_e(c)\) denotes the score assigned to pair \(c\) by the \(e\)-th evaluator according to the predefined five-point scoring rubric.

The results are reported in Table~\ref{tab:concept_quality_evaluation}, with concept-level scores provided in Appendix~\ref{appendix:concept_level_scores}. PriCBMs achieve the highest CQS of 3.71 and obtain the highest average score under each of the three evaluators. Specifically, PriCBMs obtain average scores of 3.61, 4.03, and 3.50 from GPT-5.5, Gemini-3.6-Flash, and Claude-Opus-5, respectively. Under this model-based evaluation protocol, PriCBMs receive higher average concept-quality scores than LaBo and the evaluated ablation variants on the sampled examples for overall concept quality.

\begin{center}
\begin{minipage}{1.0\linewidth}
\centering

\captionof{table}{Quantitative concept quality evaluation results.}
\label{tab:concept_quality_evaluation}

\begin{threeparttable}
\centering
{\normalsize
\setlength{\tabcolsep}{4pt}
\renewcommand{\arraystretch}{1.0}
\setlength{\aboverulesep}{0.25ex}
\setlength{\belowrulesep}{0.25ex}
\setlength{\cmidrulesep}{0.15ex}

\begin{tabular*}{0.90\linewidth}{@{\extracolsep{\fill}} lcccc @{}}
\toprule
\textbf{Method} &
\textbf{GPT-5.5} &
\textbf{G-3.6-F} &
\textbf{C-Opus-5} &
\textbf{CQS (\(\uparrow\))} \\
\midrule
LaBo        & 2.89 & 3.64 & 2.61 & 3.05 \\
w/o SDS     & 3.17 & 3.75 & 2.75 & 3.22 \\
w/o CBS     & 3.19 & 3.64 & 2.94 & 3.26 \\
w/o CS      & 3.11 & 3.64 & 2.81 & 3.19 \\
\textbf{PriCBMs}
            & \textbf{3.61} & \textbf{4.03} & \textbf{3.50} & \textbf{3.71} \\
\bottomrule
\end{tabular*}
}
\end{threeparttable}

\vspace{0.3em}

\begin{minipage}{0.90\linewidth}
\footnotesize
\justifying
SDS denotes Semantic Discriminability Selection; CBS denotes Clustering-based Selection; CS denotes Concept Selection; G-3.6-F denotes Gemini-3.6-Flash; C-Opus-5 denotes Claude-Opus-5.
\end{minipage}

\end{minipage}
\end{center}

\section{Conclusion}
\label{sec:conclusion}

This paper introduced PriCBMs for multi-label visual privacy recognition, replacing direct image-to-label prediction with an explicit concept bottleneck that guides the model to first identify category-relevant privacy cues through visual--concept matching and then predict privacy categories from the resulting concept responses. By constructing a concept bank based on target-category relevance, cross-category discriminability, and semantic diversity, PriCBMs provide category-relevant semantic representations for privacy prediction. In addition, the global-region modeling strategy combines whole-image and region-level concept responses to represent privacy-related information across spatial granularities.

Experiments on VISPR and FineViP demonstrate that PriCBMs outperform the evaluated baselines across all four multi-label metrics on both datasets under the reported settings. The ablation studies support the effectiveness of concept selection and global-region modeling, while the concept quality analysis and visualizations provide evidence of the semantic quality and relevance of the selected concepts on the 12 sampled images under the model-based evaluation protocol. These results suggest that concept-mediated modeling offers a promising way to improve multi-label visual privacy recognition while providing an explicit semantic interface for concept-level interpretation.

PriCBMs still have two main limitations. First, the current regional branch relies on fixed grid-based pooling to construct region-level visual representations. Such predefined regions may not always align well with the actual boundaries of privacy-sensitive objects, particularly for small or irregularly distributed visual cues. Future work will explore object-aware regional modeling using detection or segmentation to obtain more semantically meaningful regions for concept response computation. Second, the generalization of PriCBMs across different datasets and visual domains has not yet been fully verified. Future work will conduct cross-dataset and domain-shift evaluations and explore domain-robust concept modeling to improve generalization across different visual domains.

\section*{Data availability}

Both datasets are publicly available: VISPR at
\url{https://tribhuvanesh.github.io/vpa/}
and FineViP at
\url{https://github.com/aiprivacygroup-member1/FineViP}.

\section*{Funding}

This work was supported by the National Natural Science Foundation of China
(Grant No. 62266035).

The funder had no role in study design, data collection and analysis, manuscript preparation, or the decision to submit the article for publication.

\section*{Competing interests}

The authors declare no competing interests.

\section*{Use of AI tools}

During the preparation of this manuscript, the authors used ChatGPT
(OpenAI) to improve the language and readability of the text.
After using this tool, the authors reviewed, verified, and edited the
content as needed. The authors take full responsibility for the content
of the publication.

\section*{Author contributions}

\textbf{Shenglei Pei}: Conceptualization, Methodology, Validation,
Formal analysis, Investigation, Writing -- original draft,
Writing -- review \& editing, Funding acquisition.
\textbf{Haotian Sun}: Conceptualization, Methodology, Validation,
Formal analysis, Investigation, Writing -- original draft,
Writing -- review \& editing.
\textbf{Pengfei Li}: Conceptualization, Methodology, Supervision,
Writing -- review \& editing.
\textbf{Bingxu Lu}: Conceptualization, Methodology, Supervision,
Writing -- review \& editing.
\textbf{Wenlong Yu}: Conceptualization, Methodology, Supervision,
Writing -- review \& editing.

\clearpage
\appendix

\renewcommand{\thesection}{\Alph{section}}

\renewcommand{\thefigure}{\Alph{section}.\arabic{figure}}
\renewcommand{\thetable}{\Alph{section}.\arabic{table}}
\renewcommand{\theequation}{\Alph{section}.\arabic{equation}}

\makeatletter
\@addtoreset{figure}{section}
\@addtoreset{table}{section}
\@addtoreset{equation}{section}
\makeatother

\section{Model Configuration Analysis}
\label{appendix:configuration_analysis}

\subsection{Effect of the Number of Concepts}
\label{appendix:effect_number_concepts}

\begin{center}
\begin{minipage}{1.0\linewidth}
\centering
\captionof{table}{Test-set sensitivity analysis with different \ensuremath{K_c} settings on VISPR. The default value \ensuremath{K_c=30} is selected on the validation set before final test evaluation.}
\label{tab:kc_analysis}

{\normalsize
\setlength{\tabcolsep}{4pt}
\renewcommand{\arraystretch}{1.0}
\setlength{\aboverulesep}{0.25ex}
\setlength{\belowrulesep}{0.25ex}
\setlength{\cmidrulesep}{0.15ex}

\begin{tabular*}{0.95\linewidth}{@{\extracolsep{\fill}} lccccc @{}}
\toprule
\textbf{Dataset} &
\textbf{\ensuremath{K_c}} &
\textbf{MAP-c} &
\textbf{MAP-s} &
\textbf{Mic-F1} &
\textbf{Mac-F1} \\
\midrule
VISPR & 1  & 66.27 & 87.53 & 66.08 & 46.54 \\
VISPR & 5  & 68.13 & \textbf{87.81} & 75.14 & 54.04 \\
VISPR & 10 & \textbf{68.15} & 87.38 & 75.11 & 53.30 \\
VISPR & 30 & 67.28 & 87.30 & \textbf{82.11} & \textbf{63.51} \\
VISPR & 50 & 67.60 & 86.09 & 74.98 & 56.23 \\
\bottomrule
\end{tabular*}
}

\end{minipage}
\end{center}

\begin{center}
\begin{minipage}{1.0\linewidth}
\centering

\includegraphics[width=0.9\linewidth]{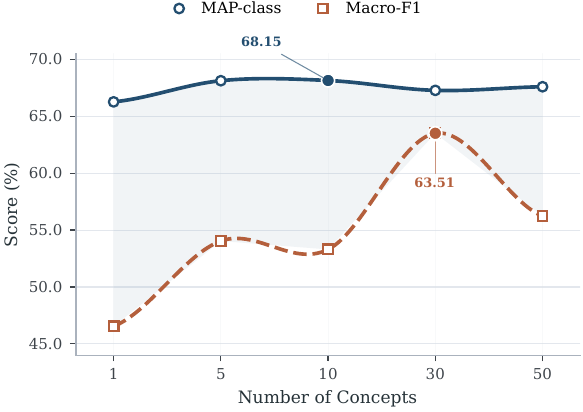}
\captionof{figure}{Test-set performance trends under different numbers of retained concepts per category on VISPR.}
\label{fig:number_concepts}

\end{minipage}
\end{center}

We evaluate the sensitivity of PriCBMs to the number of retained concepts per class, denoted as \ensuremath{K_c}, on VISPR. The default value \ensuremath{K_c=30} is selected based on the overall validation-set performance across the four evaluation metrics before final test evaluation. After fixing this validation-selected configuration, we report the corresponding test-set results under different \ensuremath{K_c} settings in Table~\ref{tab:kc_analysis} and Fig.~\ref{fig:number_concepts}. These results should be interpreted only as additional test-set sensitivity analysis for robustness assessment rather than hyperparameter-selection evidence.

All model configurations used for the final evaluation are selected on the validation set and fixed before test evaluation. In this appendix, we report additional test-set sensitivity analyses to examine how PriCBMs behave under different configuration choices in practice. These results are used only for post-hoc robustness analysis to more comprehensively assess stability under alternative settings and are not used for hyperparameter selection.

As shown in Table~\ref{tab:kc_analysis}, different values of \ensuremath{K_c} lead to different trade-offs across evaluation metrics. Smaller concept banks obtain slightly higher MAP-class or MAP-sample, whereas \ensuremath{K_c=30} achieves the strongest F1-based performance on the test set. This suggests that retaining more concepts can benefit F1-based multi-label prediction, while an excessively large concept bank may introduce redundant or less discriminative concepts. Since the default value \ensuremath{K_c=30} is selected on the validation set before test evaluation, the results in Table~\ref{tab:kc_analysis} are reported only as post-hoc sensitivity analysis for additional robustness assessment.

\subsection{Effect of the Number of Regions}
\label{appendix:effect_number_regions}

\begin{center}
\begin{minipage}{1.0\linewidth}
\centering
\captionof{table}{Test-set sensitivity analysis with different numbers of regions on VISPR. The default setting of 9 regions is selected on the validation set before final test evaluation.}
\label{tab:region_number}

{\normalsize
\setlength{\tabcolsep}{4pt}
\renewcommand{\arraystretch}{1.0}
\setlength{\aboverulesep}{0.25ex}
\setlength{\belowrulesep}{0.25ex}
\setlength{\cmidrulesep}{0.15ex}

\begin{tabular*}{0.95\linewidth}{@{\extracolsep{\fill}} lcccc @{}}
\toprule
\textbf{Region} &
\textbf{MAP-c} &
\textbf{MAP-s} &
\textbf{Mic-F1} &
\textbf{Mac-F1} \\
\midrule
2  & 67.60 & \textbf{87.77} & 80.50 & 61.71 \\
4  & \textbf{68.96} & 87.12 & 82.03 & 63.37 \\
9  & 67.28 & 87.30 & 82.11 & \textbf{63.51} \\
16 & 67.58 & 87.20 & \textbf{82.14} & 62.36 \\
24 & 67.23 & 86.77 & 81.47 & 61.84 \\
32 & 66.51 & 87.26 & 81.05 & 60.07 \\
\bottomrule
\end{tabular*}
}

\vspace{0.8em}

\includegraphics[width=0.9\linewidth]{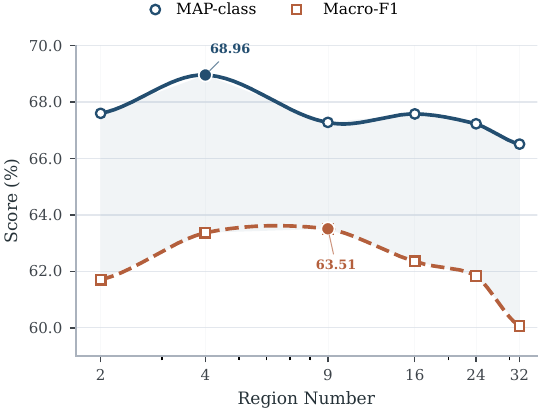}
\captionof{figure}{Test-set performance trends under different region number settings on VISPR.}
\label{fig:region_number}

\end{minipage}
\end{center}

\FloatBarrier

We further evaluate the sensitivity of PriCBMs to the number of local regions on VISPR under different regional granularity settings. The evaluated settings \(R\in\{2,4,9,16,24,32\}\) correspond to \(R_h\times R_w\) grids of \(1\times2\), \(2\times2\), \(3\times3\), \(4\times4\), \(4\times6\), and \(4\times8\), respectively. The default setting of 9 regions is selected based on the overall validation-set performance across the four evaluation metrics and fixed before final test evaluation. After this validation-based selection, we report the corresponding test-set results under different region numbers in Table~\ref{tab:region_number} and Fig.~\ref{fig:region_number}. These results are used only to analyze the effect of regional granularity and are not used to select the final region configuration. By varying the number of regions, we examine how different spatial partitioning granularities affect the balance between localized privacy evidence, overall prediction stability, and the resulting classification performance under different region settings.

As shown in Table~\ref{tab:region_number}, performance varies with the number of local regions. Different region settings achieve the best results on different metrics: 4 regions obtain the highest MAP-class, 2 regions achieve the highest MAP-sample, 16 regions obtain the highest Micro-F1, and 9 regions achieve the highest Macro-F1. This indicates that different regional granularities lead to different trade-offs across the evaluation metrics. Using fewer regions provides coarser spatial representations, whereas increasing the number of regions allows privacy evidence to be modeled at a more localized level with finer spatial partitioning. However, increasing the number of regions does not consistently improve classification performance. Since the final setting of 9 regions is selected on the validation set before test evaluation, the test-set results in Table~\ref{tab:region_number} are interpreted as configuration sensitivity for additional post-hoc analysis rather than model-selection evidence.

\section{Binary Classification Comparison}
\label{appendix:binary_classification}

Since most existing privacy classification studies formulate the task as a privacy-sensitive/safe binary classification problem, we construct a task-specific binary variant, termed PriCBMs-Binary, and compare it with representative interpretable methods, including PEAK~\cite{ayci2023peak}, Priv$\times$ITM~\cite{baia2024image}, and PrivLEX~\cite{baranouskaya2026privlex}. PriCBMs-Binary is trained independently under the Safe/Public-versus-Privacy label space rather than obtained by collapsing the multi-label predictions.  Specifically, PriCBMs-Binary reuses the original privacy concept bank from the multi-label model and introduces a Safe/Public concept set of the same size as the original privacy concept bank. The multi-label concept-to-class head is replaced with a binary head, while the remaining components are unchanged. The results are reported in Table~\ref{tab:binary_comparison}.

As shown in Table~\ref{tab:binary_comparison}, PriCBMs-Binary achieves the highest performance on both VISPR and FineViP in terms of accuracy, balanced accuracy, and Macro-F1. It obtains 91.38\%, 91.01\%, and 90.86\% on VISPR, and 98.34\%, 92.22\%, and 89.66\% on FineViP, respectively. PriCBMs-Binary also achieves the highest F1-scores for both the Safe/Public and Privacy classes on the two datasets, indicating more balanced binary classification performance across the two classes.

Several methods show relatively low \textit{Public}-class performance on the imbalanced FineViP test split. In the FineViP test split used for binary evaluation, \textit{Public} accounts for 3.92\% of the samples, whereas the remaining 96.08\% belong to the \textit{Privacy} class. Under such a skewed distribution, high overall accuracy can coexist with poor recall for the minority \textit{Public} class. For example, PEAK achieves 96.11\% accuracy but only 1.92\% recall for the \textit{Public} class. Therefore, balanced accuracy and Macro-F1 provide a more informative evaluation than overall accuracy under this imbalanced setting.

\begin{center}
\begin{minipage}{1.0\linewidth}
\centering

\captionof{table}{Binary classification comparison results on the VISPR and FineViP datasets.}
\label{tab:binary_comparison}

\begin{threeparttable}
\centering
{\fontsize{10pt}{12pt}\selectfont
\setlength{\tabcolsep}{1.6pt}
\renewcommand{\arraystretch}{1.0}
\setlength{\aboverulesep}{0.25ex}
\setlength{\belowrulesep}{0.25ex}
\setlength{\cmidrulesep}{0.15ex}

\begin{tabular*}{\linewidth}{@{\extracolsep{\fill}} llccccccccc @{}}
\toprule
\multirow{2}{*}{\textbf{Dataset}} &
\multirow{2}{*}{\textbf{Model}} &
\multirow{2}{*}{\textbf{ACC}} &
\multirow{2}{*}{\textbf{BA}} &
\multirow{2}{*}{\textbf{F1-m}} &
\multicolumn{3}{c}{\textbf{Public/Safe}} &
\multicolumn{3}{c}{\textbf{Privacy}} \\
\cmidrule(lr){6-8}
\cmidrule(lr){9-11}
& & & & &
\textbf{P} & \textbf{R} & \textbf{F1} &
\textbf{P} & \textbf{R} & \textbf{F1} \\
\midrule

\multirow{4}{*}{VISPR}
& PEAK            & 85.47 & 83.58 & 84.20 & 83.86 & 75.96 & 79.71 & 86.30 & 91.20 & 88.68 \\
& Priv{$\times$}ITM & 87.16 & 86.33 & 86.32 & 82.88 & 82.97 & 82.93 & 89.74 & 89.69 & 89.71 \\
& PrivLEX         & 88.84 & 88.11 & 88.10 & 85.14 & 85.16 & 85.15 & 91.07 & 91.05 & 91.06 \\
& PriCBMs-B
& \textbf{91.38} & \textbf{91.01} & \textbf{90.86}
& \textbf{87.88} & \textbf{89.44} & \textbf{88.65}
& \textbf{93.58} & \textbf{92.57} & \textbf{93.07} \\
\midrule

\multirow{4}{*}{FineViP}
& PEAK            & 96.11 & 50.94 & 50.88 & 66.67 & 1.92 & 3.74 & 96.15 & \textbf{99.96} & 98.02 \\
& Priv{$\times$}ITM & 96.76 & 69.72 & 73.87 & 63.64 & 40.38 & 49.41 & 97.60 & 99.06 & 98.32 \\
& PrivLEX         & 96.72 & 64.63 & 69.96 & 68.89 & 29.81 & 41.61 & 97.20 & 99.45 & 98.31 \\
& PriCBMs-B
& \textbf{98.34} & \textbf{92.22} & \textbf{89.66}
& \textbf{75.42} & \textbf{85.58} & \textbf{80.18}
& \textbf{99.41} & 98.86 & \textbf{99.13} \\
\bottomrule
\end{tabular*}
}

\vspace{0.5mm}

\begin{tablenotes}[flushleft]
\footnotesize
\item \parbox[t]{\dimexpr\linewidth-3pt\relax}{\raggedright
PriCBMs-B: PriCBMs-Binary;
ACC: accuracy;
BA: balanced accuracy;
F1-m: Macro-F1;
P/R/F1: precision/recall/F1-score.
}
\end{tablenotes}

\end{threeparttable}
\end{minipage}
\end{center}

\FloatBarrier

\section{Complete Visualization of Selected Concepts}
\label{appendix:selected_concepts_full}

To demonstrate the interpretability of PriCBMs, we provide the visualization results of selected concepts in Fig.~\ref{fig:xai_concept_full}. For the sampled images, the selected concepts are generally consistent with the visible image content and privacy-related semantics, supporting their qualitative semantic relevance on these examples.

\begin{center}
\begin{minipage}{1.0\linewidth}
\centering
\captionof{figure}{Complete visualization of selected concepts for VISPR images~\cite{ref3}. For each image, the top three are ranked by combined global and regional responses to illustrate image-concept associations.}
\label{fig:xai_concept_full}

\vspace{0.2em}
\setstretch{1}
\includegraphics[
    width=\linewidth,
    height=0.82\textheight,
    keepaspectratio
]{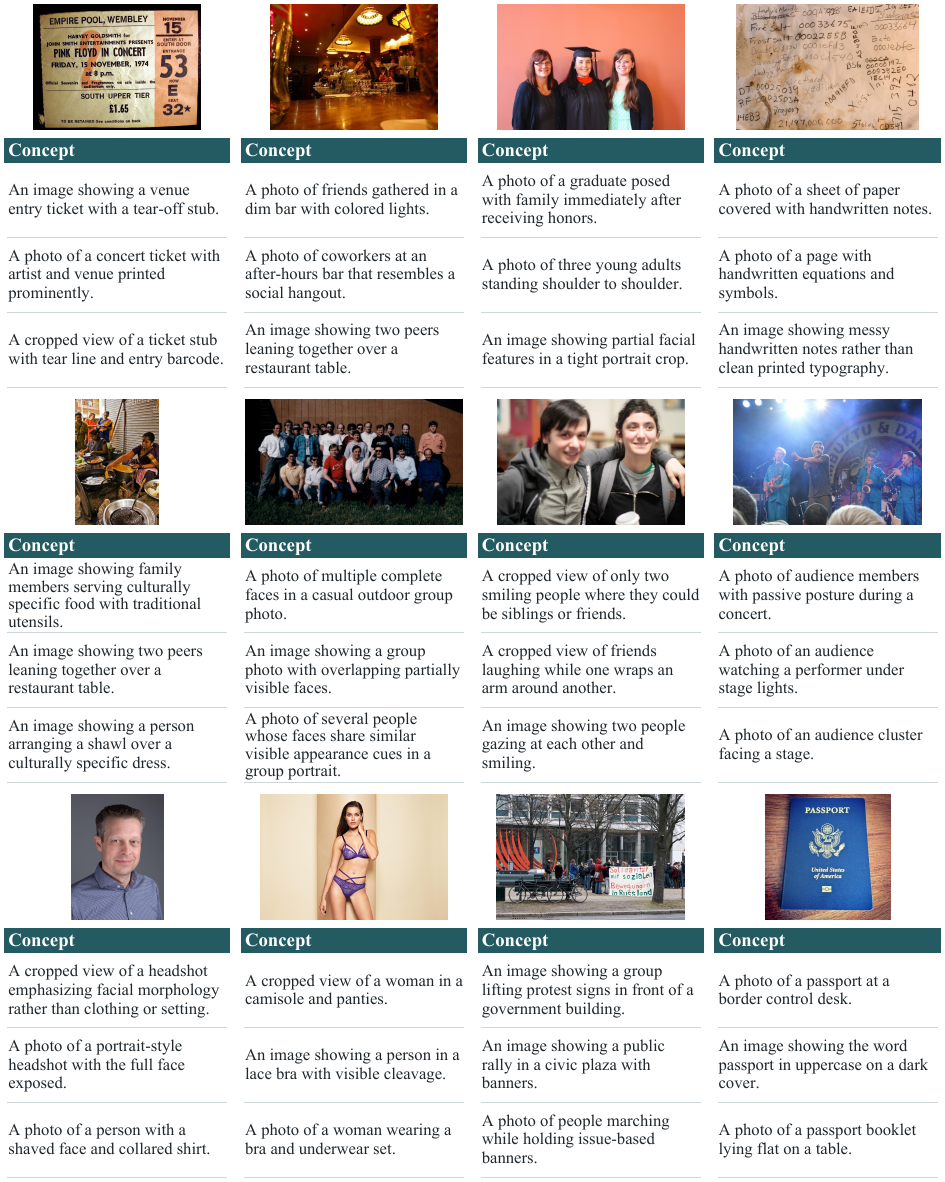}
\end{minipage}
\end{center}

\FloatBarrier

\section{Detailed Concept-level Scores}
\label{appendix:concept_level_scores}

This appendix provides the detailed concept-level quality evaluation results corresponding to the quantitative analysis in Section~\ref{sec:interpretability_comparison}. For each sampled image, the top three concepts selected by each compared method are shown together with the scores assigned by three separately queried large vision-language model evaluators, namely GPT-5.5, Gemini-3.6-Flash, and Claude-Opus-5.

The compared methods include LaBo, w/o Semantic Discriminability Selection, w/o Clustering-based Selection, w/o Concept Selection, and the full PriCBMs. Each concept is scored on a five-point scale according to the rubric defined in Table~\ref{tab:concept_quality_rubric}. A higher score indicates that the concept is more visually grounded, semantically accurate, and explanatory for the corresponding image.

\begin{center}
\begin{minipage}{1.0\linewidth}
\centering

\captionof{table}{Scoring rubric for concept quality evaluation.}
\label{tab:concept_quality_rubric}

{\fontsize{10pt}{12pt}\selectfont
\setlength{\tabcolsep}{4pt}
\renewcommand{\arraystretch}{1.08}
\setlength{\aboverulesep}{0.35ex}
\setlength{\belowrulesep}{0.35ex}

\begin{tabular*}{0.95\linewidth}{@{\extracolsep{\fill}} c p{0.78\linewidth} @{}}
\toprule
\textbf{Score} & \textbf{Meaning} \\
\midrule
1 & Clearly wrong, hallucinated, or unrelated to the image. \\
2 & Partially related, but with obvious inaccuracies or hallucinated details. \\
3 & Basically related, but generic or partially uncertain. \\
4 & Relevant, clear, and visually verifiable. \\
5 & Highly accurate, specific, and strongly explanatory. \\
\bottomrule
\end{tabular*}
}

\end{minipage}
\end{center}

\FloatBarrier

\begin{figure}[!p]
\centering
\includegraphics[width=1.0\linewidth,height=0.84\textheight,keepaspectratio]{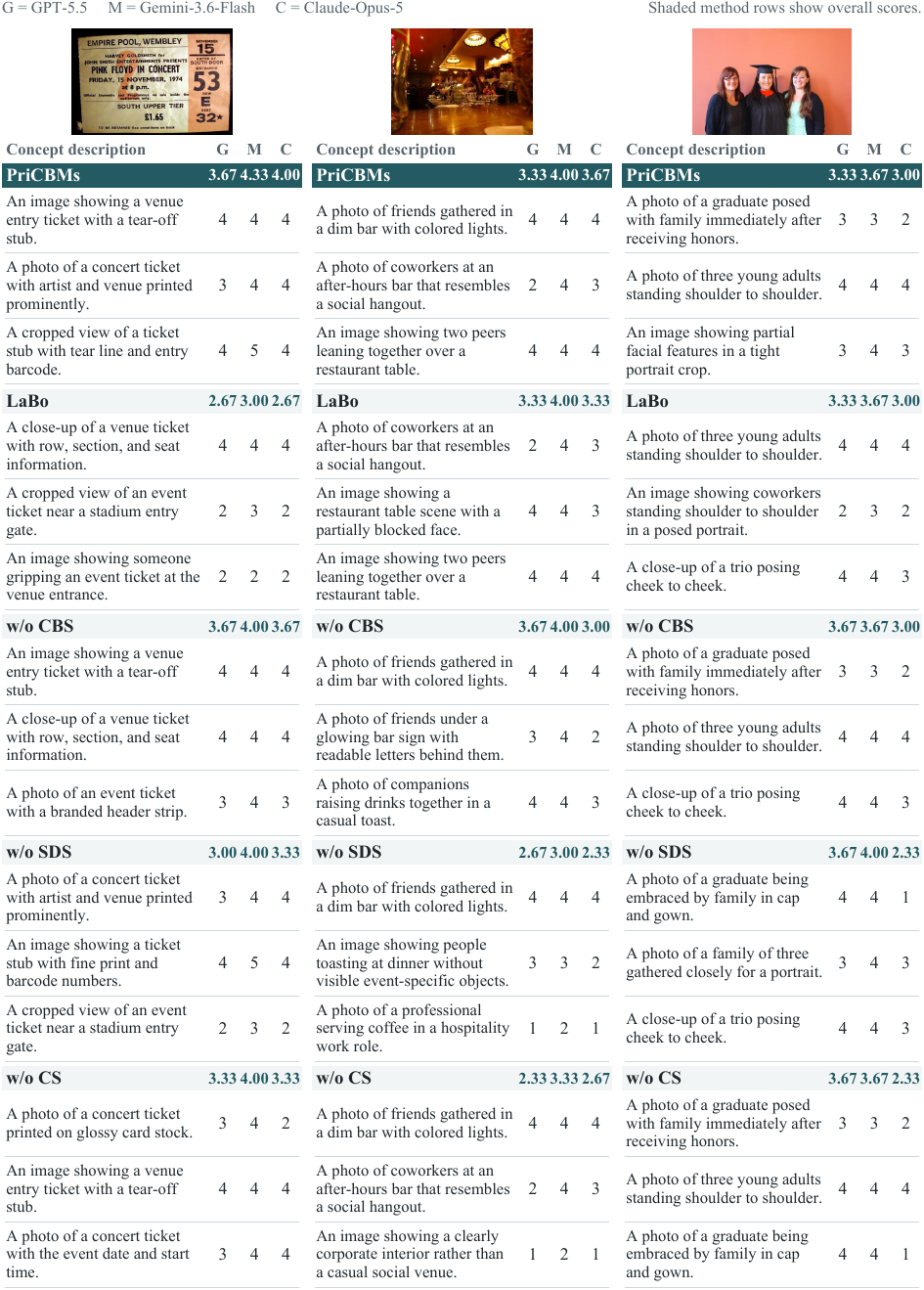}
\caption{Detailed concept-level quality scores for sampled VISPR examples~\cite{ref3}, part 1. For each image, the top three concepts of different methods and their evaluator scores are shown.}
\label{fig:concept_score_1}
\end{figure}

\begin{figure}[!p]
\centering
\includegraphics[width=1.0\linewidth,height=0.84\textheight,keepaspectratio]{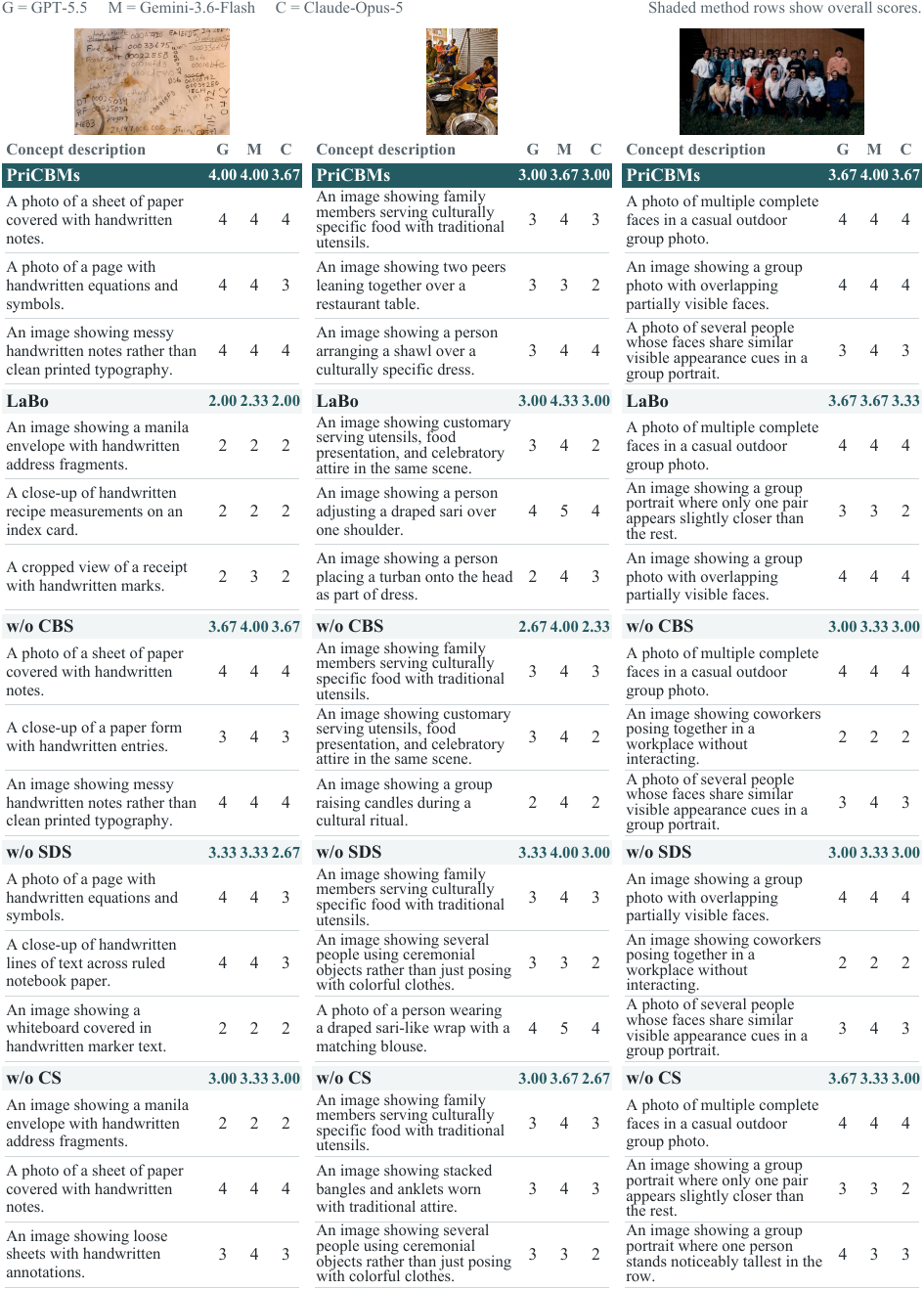}
\caption{Detailed concept-level quality scores for sampled VISPR examples~\cite{ref3}, part 2. For each image, the top three concepts of different methods and their evaluator scores are shown.}
\label{fig:concept_score_2}
\end{figure}

\begin{figure}[!p]
\centering
\includegraphics[width=1.0\linewidth,height=0.84\textheight,keepaspectratio]{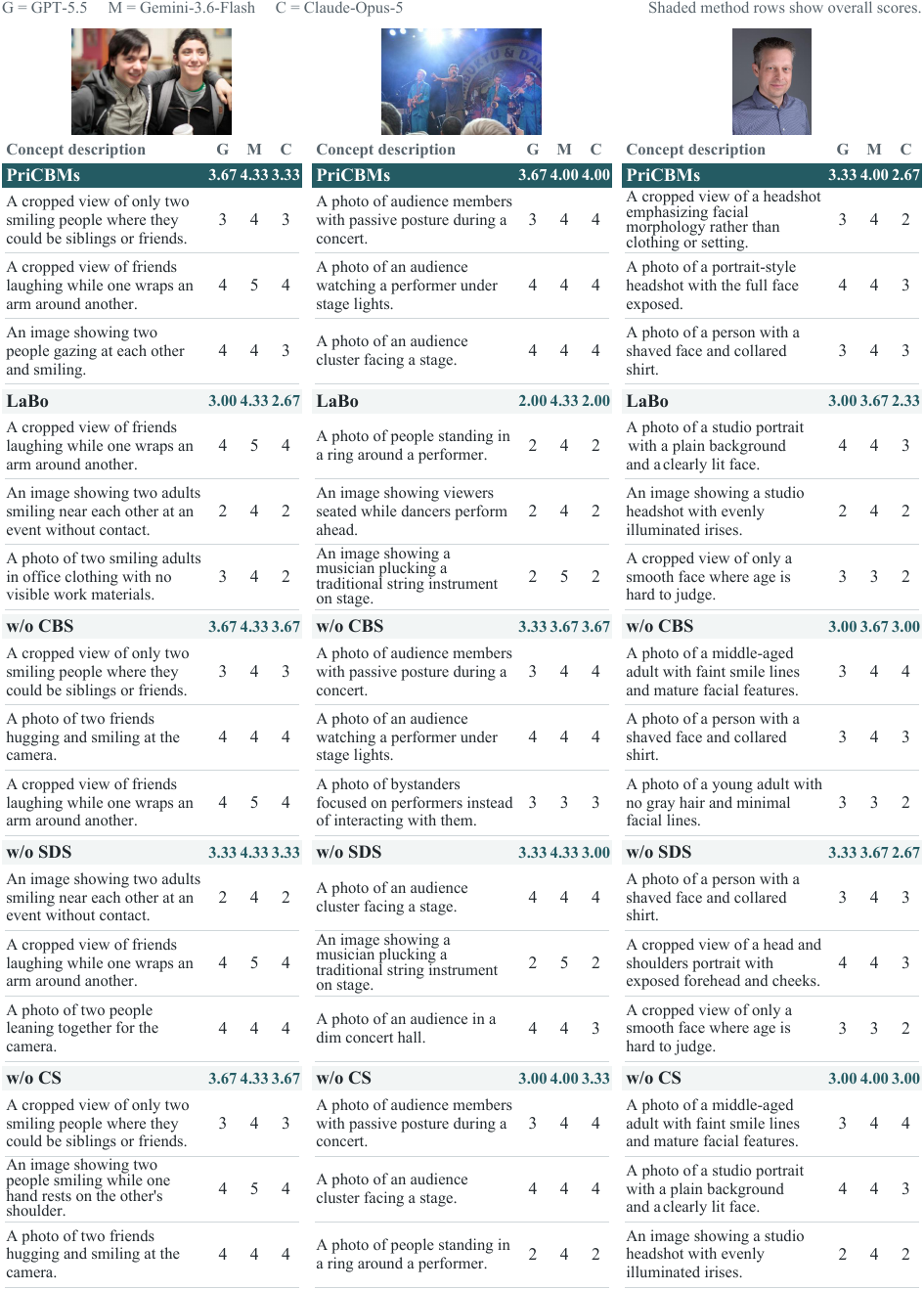}
\caption{Detailed concept-level quality scores for sampled VISPR examples~\cite{ref3}, part 3. For each image, the top three concepts of different methods and their evaluator scores are shown.}
\label{fig:concept_score_3}
\end{figure}

\begin{figure}[!p]
\centering
\includegraphics[width=1.0\linewidth,height=0.84\textheight,keepaspectratio]{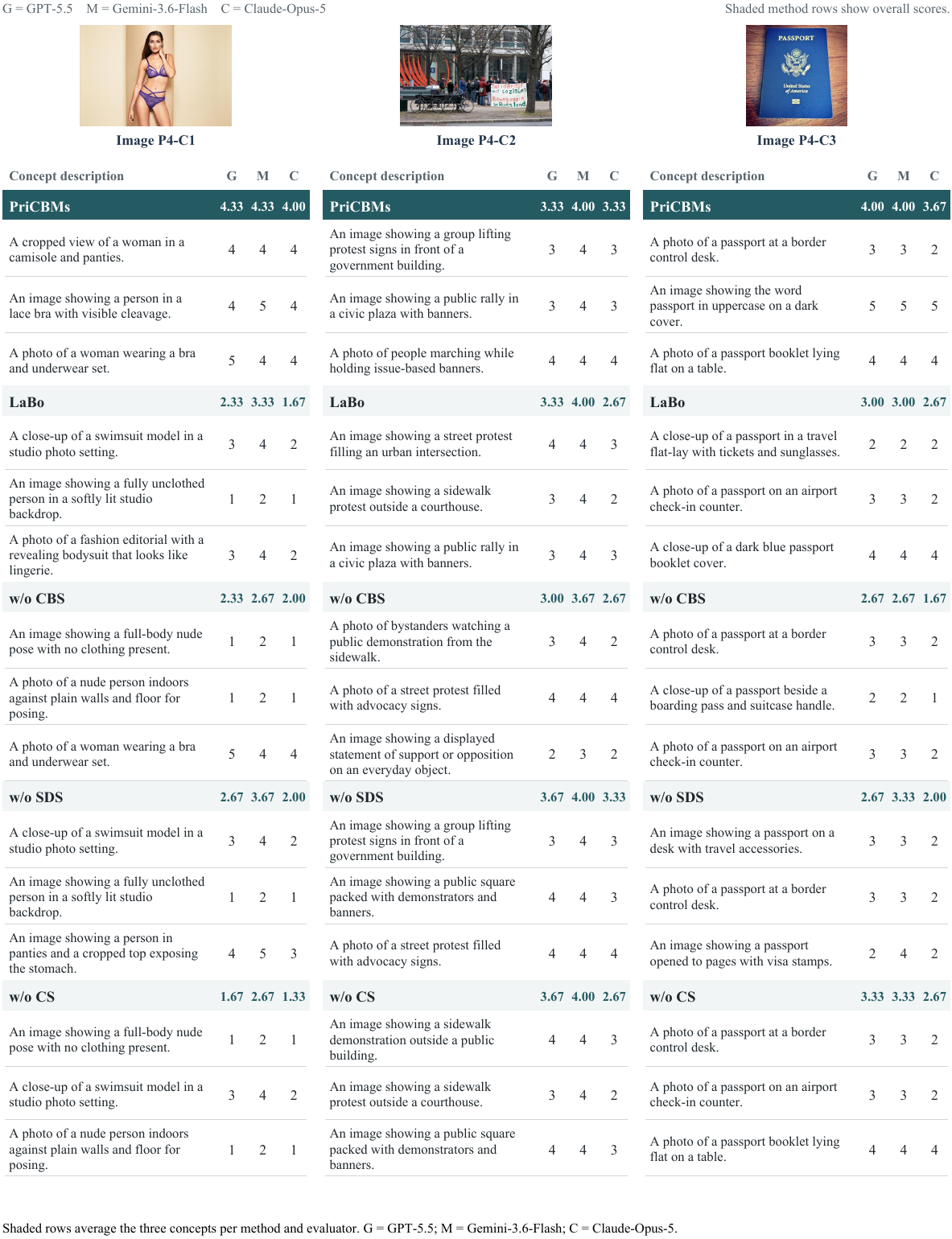}
\caption{Detailed concept-level quality scores for sampled VISPR examples~\cite{ref3}, part 4. For each image, the top three concepts of different methods and their evaluator scores are shown.}
\label{fig:concept_score_4}
\end{figure}

The detailed scoring results are shown in Figs.~\ref{fig:concept_score_1}--\ref{fig:concept_score_4}.

\FloatBarrier

\end{document}